\documentclass[preprint,1p,times]{elsarticle}

\usepackage{amsmath}
\usepackage{amssymb}
\usepackage{hyperref}
\usepackage{graphicx}
\usepackage{subfigure}
\usepackage{setspace}
\usepackage{bm}
\usepackage[normalem]{ulem}
\usepackage{color}
\usepackage{lineno,hyperref}
\usepackage{stfloats}
\usepackage[justification=centering]{caption}
\usepackage{setspace} 
\usepackage{setspace} 
\usepackage{array}
\modulolinenumbers[1]
\usepackage{makecell}
\usepackage{changes}
\usepackage{mdframed}
\usepackage{multicol}
\usepackage{multirow}
\usepackage{framed} % Framing content
\usepackage{multicol} % Multiple columns environment
\usepackage{booktabs}
\usepackage{threeparttable}
\usepackage{adjustbox}
\usepackage[T1]{fontenc}
\usepackage[tableposition=top]{caption}
\usepackage{tabularx}
\usepackage{soul}
\usepackage{url}
\usepackage{nomencl} % Nomenclature package
\biboptions{numbers,sort&compress}
\makeatletter
\def\ps@pprintTitle{%
	\let\@oddhead\@empty
	\let\@evenhead\@empty
	\let\@oddfoot\@empty
	\let\@evenfoot\@oddfoot
}
\makeatother
\begin{document}

\begin{frontmatter}

	\title{A Self-attention Knowledge Domain Adaptation Network for Commercial Lithium-ion Batteries State-of-Health Estimation under Shallow Cycles}
	\author[main]{Xin Chen\corref{cor1}}
	\author[main]{Yuwen Qin}
	\author[main]{Weidong Zhao}
	\author[second]{Qiming Yang}
	\author[main]{Ningbo Cai}
	\author[second]{Kai Wu}
	\cortext[cor1]{E-mail: xin.chen.nj@xjtu.edu.cn}
	\address[main]{Center of Nanomaterials for Renewable Energy, State Key Laboratory of Electrical Insulation and Power Equipment, School of Electrical Engineering, Xi'an Jiaotong University, Xi'an 710054, Shaanxi, China.}
	\address[second]{State Key Laboratory of Electrical Insulation and Power Equipment, School of Electrical Engineering, Xi'an Jiaotong University, Xi'an 710054, Shaanxi, China.}

	\begin{abstract}
	Accurate state-of-health (SOH) estimation is critical to guarantee the safety, efficiency and reliability of battery-powered applications. Most SOH estimation methods focus on the 0-100\% full state-of-charge (SOC) range that has similar distributions. However, the batteries in real-world applications usually work in the partial SOC range under shallow-cycle conditions and follow different degradation profiles with no labeled data available, thus making SOH estimation challenging. To estimate shallow-cycle battery SOH, a novel unsupervised deep transfer learning method is proposed to bridge different domains using self-attention distillation module and multi-kernel maximum mean discrepancy technique. The proposed method automatically extracts domain-variant features from charge curves to transfer knowledge from the large-scale labeled full cycles to the unlabeled shallow cycles. The CALCE and SNL battery datasets are employed to verify the effectiveness of the proposed method to estimate the battery SOH for different SOC ranges, temperatures, and discharge rates. The proposed method achieves a root-mean-square error within 2\% and outperforms other transfer learning methods for different SOC ranges. When applied to batteries with different operating conditions and from different manufacturers, the proposed method still exhibits superior SOH estimation performance. The proposed method is the first attempt at accurately estimating battery SOH under shallow-cycle conditions without needing a full-cycle characteristic test.
	\end{abstract}

	%Graphical abstract
%	\begin{graphicalabstract}
%		\resizebox{1\textwidth}{!}{%
%			\includegraphics{figures/structure_2.pdf}
%		}
%	\end{graphicalabstract}

	%Research highlights
	% \begin{highlights}
	% \item A novel domain adaptation model is developed for shallow-cycle SOH estimation.
	
	% \item The self-attention distillation module is applied to extract domain-invariant features.
	
	% \item Accurate SOH estimation of batteries in shallow cycles are achieved.
	
	% \item The proposed model shows robustness against different SOC ranges and operating conditions.
	
	% \item The proposed model is insensitive to battery manufacturers and shallow-cycle degradation.
 	
	% \end{highlights}

	% \begin{keyword}
	% Li-ion batteries\sep Shallow cycles \sep SOH estimation \sep Unsupervised deep transfer learning \sep Attention mechanism 
	% \end{keyword}

\end{frontmatter}

% \maketitle
% \linenumbers

\section{Introduction}
Rechargeable lithium-ion (Li-ion) batteries have emerged as the primary energy storage solution for mobile devices and smart grids\cite{ng2020predicting,roman2022design,yang2021review}. With repeated charge-discharge cycles, the battery capacity continuously degrades due to irreversible physical, chemical, and mechanical changes\cite{hu2020battery}. Battery state-of-health (SOH) is generally used to quantify the degradation degree of a battery and is defined as the ratio of the present maximum capacity with respect to the initial maximum capacity\cite{roman2021machine}. The accurate battery SOH estimation is crucial but remains challenging for dynamic operating conditions and random user behavior. Conventionally, the present maximum capacity is typically calculated from a full discharge curve after a full charge according to its definition, but this is not feasible for random charge and discharge conditions with flexible starting and ending points in real-world applications\cite{tian2021deep,bockrath2023state,jenu2022state}. During the battery degradation process, few recorded curves are full discharge curves in the battery-management system (BMS), and most are partial curves. Therefore, it is imperative to develop accurate methods for estimating the battery SOH based on partial charge/discharge curves\cite{tian2022flexible,wei2022multistage}.

In recent years, numerous methods have been proposed to estimate battery SOH, including model-based and data-driven methods. Model-based methods mainly adopt equivalent circuit models or electrochemical models to simulate the degradation trajectory of the battery SOH. Due to more complex degradation paths and limited sensor information available in the partial charge/discharge conditions, it can be challenging to accurately estimate battery SOH using physical models with a clear electrochemical mechanism\cite{vennam2022survey,xu2022co,liu2022online,li2022novel}. In contrast, data-driven methods with strong nonlinear fitting capabilities can directly map partial charge/discharge curves to SOH, offering a promising alternative to model-based methods\cite{che2022data,ruan2023artificial}.

Data-driven methods can be further categorized into feature-based methods and sequence-based methods\cite{deng2022battery}. Feature-based methods rely on constructing handcrafted features from charge/ discharge curves and then establishing a mapping relationship between features and battery SOH by traditional machine learning algorithms, such as support vector machines\cite{feng2019online}, random forest\cite{roman2021machine}, artificial neural network\cite{luo2022online}, and gaussian process regression\cite{jin2023state}. Among various feature-extraction techniques, the incremental capacity and differential voltage (IC/DV) analysis are widely used in the literature\cite{wang2022health,liu2022comparative,tian2023capacity,brunetaud2023non,pan2022integration}. The evolution of IC/DA curves is capable of characterizing the degradation process of batteries, where the heights, positions, and shapes of the IC/DV peaks are highly related to battery SOH. Nevertheless, feature-based methods usually require a sufficient voltage range and are sensitive to the noise in charge/discharge curves. Sequence-based methods automatically extract degradation-related features from raw partial charge/discharge curves, eliminating the need for laborious feature engineering and enabling greater flexibility across different operating conditions\cite{sui2021review}. Sequence-based deep learning models such as convolutional neural networks (CNN)\cite{ruan2023artificial,fan2023battery,gu2023novel} and long and short-term memory networks (LSTM)\cite{chen2022novel,kim2022novel,ardeshiri2022multivariate} have shown promising results in performing end-to-end estimation from partial charge/discharge curves to battery SOH. However, it is worth noting that all these methods are based on a key assumption that the batteries operate in full cycles. In real-world scenarios where there is a strong demand for the safety and stability of the battery, the battery often operates under shallow cycles with a specific depth of discharge (DOD) and state-of-charge (SOC) range rather than full cycles with 0\%-100\% SOC range. Thus, the effectiveness of these methods in practical applications remains to be verified.

\begin{figure}[!h]
	\centering
	\includegraphics[width=1\textwidth]{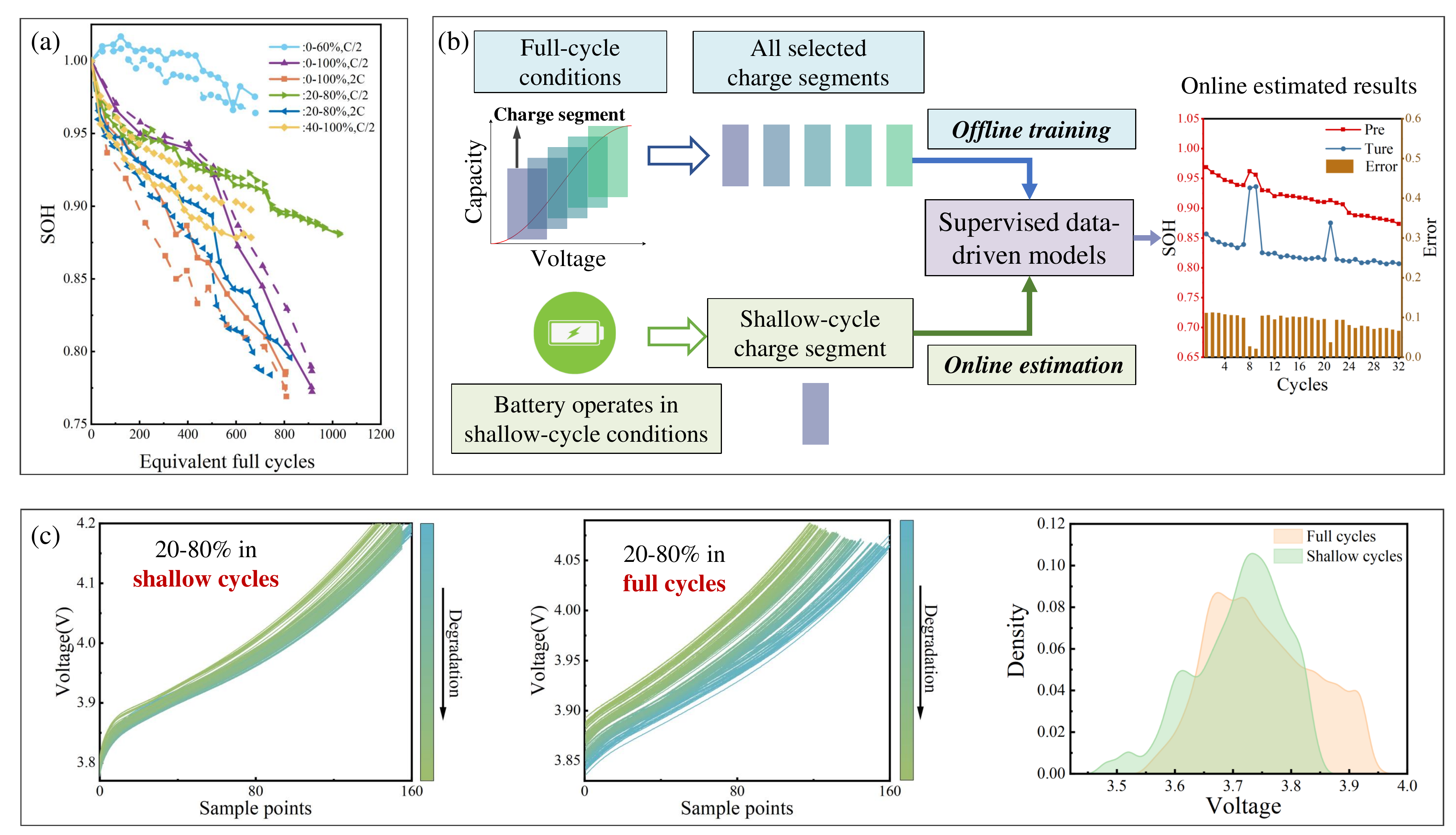}
	\caption{Poor estimation performance of battery in shallow cycles using the existing supervised data-driven models trained in full cycles. \textbf{a}, the batteries SOH degradation curves of the CALCE dataset under different SOC ranges. \textbf{b}, schematic diagram of the shallow-cycle battery SOH estimation based on partial charge curves in full cycles. \textbf{c}, the evolution of the charge process with cycles and probability distribution for batteries under different SOC ranges.}
	\label{direct_estimation}
\end{figure}

Numerous studies have proved that the DOD and mean SOC can have a significant impact on battery degradation, leading to different degradation paths\cite{pan2022integration,li2022towards,saxena2016cycle,preger2020degradation}, as shown in Fig.~\ref{direct_estimation} (a). When the charge curves in shallow cycles are put into the models that are trained with partial charge curves in full cycles (see Fig.~\ref{direct_estimation} (b)), the results show that there is a significant error between the estimated and true SOH under the shallow-cycle condition. This phenomenon arises from the domain discrepancy, which is the gap in data distribution between shallow cycles and full cycles (see Fig.~\ref{direct_estimation} (c)). Hence, it is not feasible to extrapolate the model directly from full cycles to shallow cycles. In addition, a more challenging issue is the difficulty in collecting real-time battery SOH labels. The current technique to estimate the battery SOH under shallow-cycle conditions remains a time-consuming and labor-intensive task because it demands offline measurement using complex physical and electrochemical approaches\cite{xu2022quantification, jones2022impedance,you2023situ,hou2022estimation}. As a result, the SOH of each shallow cycle is unknown in the battery historical data under shallow-cycle conditions, rendering data-driven methods that rely on supervised learning not applicable to the shallow-cycle SOH estimation. It is demanding work for the development of a novel data-driven framework for the real-time online estimation of battery SOH under shallow-cycle conditions.
\captionsetup[figure]{justification=raggedright}

Unsupervised deep transfer learning (UDTL) is usually applied to address the domain discrepancy problem\cite{zhao2021applications,zhao2020review}. However, despite variations in chemical materials, nominal capacities, and operating conditions, batteries conform to the same physical laws and degradation trends\cite{tian2021deep}. Accordingly, the UDTL is capable of learning domain-invariant features to transfer knowledge from a large-scale labeled source domain to a new unlabeled target domain. In this paper, we propose a novel UDTL-based model, named self-attention knowledge domain adaptation network (SKDAN), for estimating battery SOH under shallow-cycle conditions. As the discharge process of a battery is time-varying while the charge process is relatively fixed in real applications\cite{zhu2022data,deng2021data}, the SKDAN model is employed to automatically extract degradation-related features from the charge curves and map them to the battery SOH. The main contributions of this paper are summarized as follows:

\begin{itemize}
	\item Battery degradation curves have a substantial difference between shallow-cycle and full-cycle conditions. This paper is the first effort to study how to estimate battery SOH without resorting to additional calibrated experiments but using only unlabeled historical charge data under shallow-cycle conditions.
	
	\item An effective UDTL-based method named SKDAN is developed to bridge different datasets and operating conditions with distribution discrepancy. In the SKDAN model, the self-attention distillation (SAD) module efficiently extracts the degradation-related features with excellent domain-invariance. The multi-kernel maximum mean discrepancy (MK-MMD) is employed to align the distribution of the degradation-related features in the subspace, significantly enhancing the generalization performance of the model.
	
	\item The proposed SKDAN is verified with two commercial battery datasets operating at different temperatures, discharge rates, and SOC ranges. The results show that the proposed SKDAN achieves excellent performance in estimating the shallow-cycle SOH under different working conditions using the full-cycle source dataset. The proposed SKDAN is insensitive and robust against battery manufacturers, operating conditions, and shallow-cycle degradation.
	
\end{itemize}

The rest of the paper is organized as follows. In section \ref{model}, the structure of the proposed model is described in detail. Experiments and results on the CALCE and the SNL datasets are presented in Section \ref{experiment}. Section \ref{discussion} discusses the contribution of each part of the proposed model and the effectiveness of feature extraction. Finally, the conclusion is given in \ref{conclusion}.

\section{Methodology}\label{model}
In this section, we develop a SKDAN model to estimate battery SOH across domains with different SOC ranges and operation conditions. We first introduce the formulation and notation of UDTL problem, and then provide a detailed description of the model's structure, components, and optimization objectives.

\subsection{Problem formulation}
To briefly formulate the unsupervised deep transfer learning (UDTL) problem, we introduce some basic notations.
The labeled source domain (full-cycle conditions) is defined as
$ \mathcal{D}_s=\left\{\left(\mathbf{x}_i^s, y_i^s\right)\right\}_{i=1}^{N_s}$ and $\mathbf{x}_i^s \in \mathbf{X}_s, $ where $N_s$ is the number of samples in the source domain, $\mathbf{X}_s$ is the set of all samples, and $\mathbf{x}_i^s$ and $y_i^s$ denote the $i$-th sample and the corresponding SOH, respectively. The unlabeled target domain (shallow-cycle conditions) is similarly represented as $
\mathcal{D}_t=\left\{\left(\mathbf{x}_i^t\right)\right\}_{i=1}^{N_t}$ and $ \mathbf{x}_i^t \in \mathbf{X}_t$. 
$ \mathcal{D}_s$ and $\mathcal{D}_t$ are separately sampled from two marginal probability distributions $\mathcal{P}(\mathbf{X}_s)$, $\mathcal{Q}(\mathbf{X}_t)$, where $\mathcal{P}(\mathbf{X}_s)\neq\mathcal{Q}(\mathbf{X}_t)$ due to different operating conditions. 
Since the two domains share the same learning task, this paper aims to transfer knowledge from the labeled $ \mathcal{D}_s $ to the unlabeled $ \mathcal{D}_t $ for SOH estimation. That is, our goal is to learn a function $h$ which approximates the SOH of the shallow-cycle battery directly from the charge data, i.e.$ y_i^t \approx h\left(\mathbf{x}_i^t\right)$.

\subsection{Self-attention knowledge domain adaptation network}
\begin{figure}[!h]
	\centering
	\includegraphics[width=0.9\textwidth]{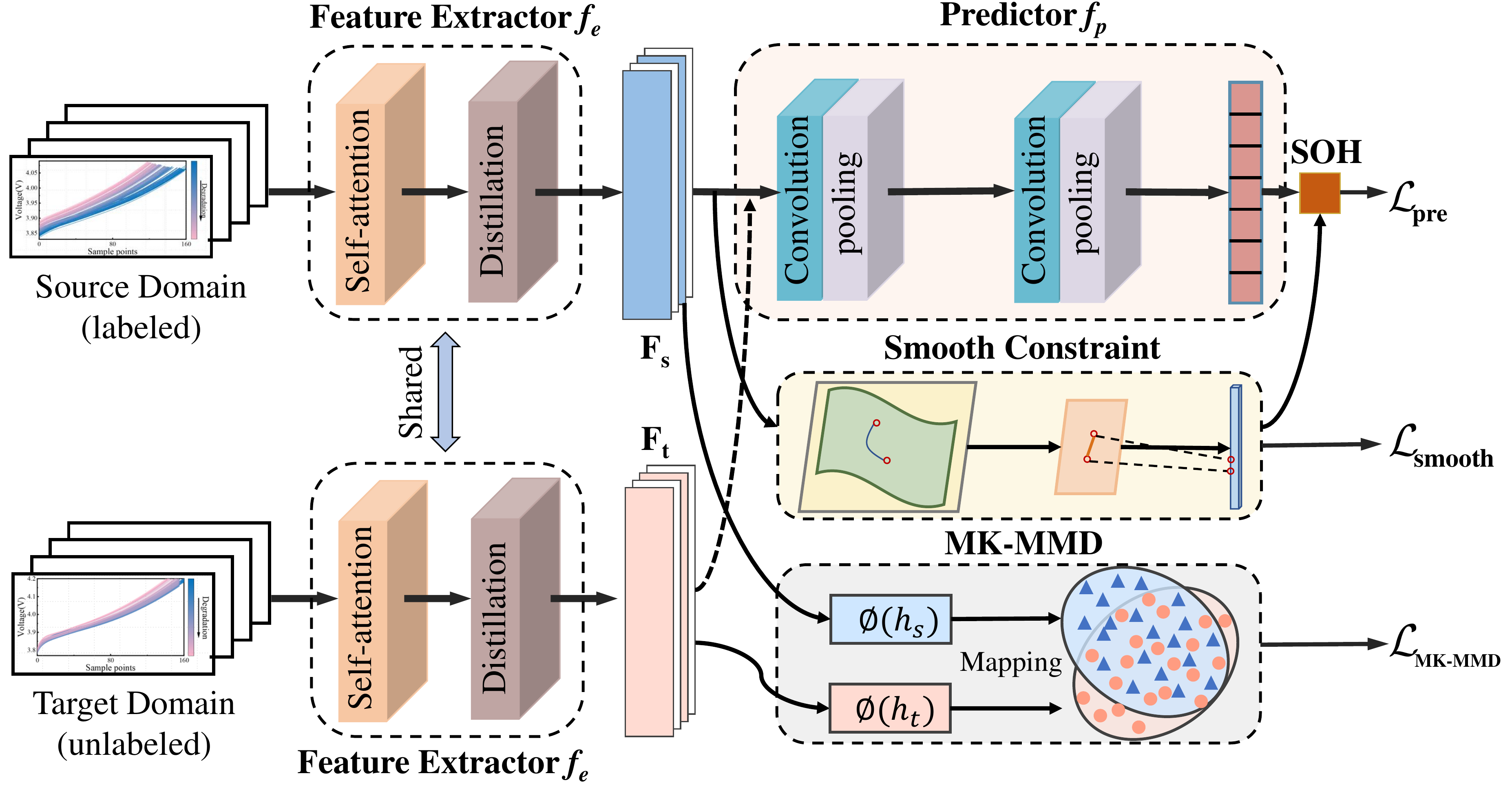}
	\caption{Architecture of the proposed SKDAN for shallow-cycle SOH estimation.}
	\label{SKDAN_structure}
\end{figure}

The SKDAN model is proposed to deal with the problem of domain discrepancy between shallow cycles and full cycles. As shown in Fig.~\ref{SKDAN_structure}, the SKDAN model consists of three main parts, SAD feature extractor, CNN predictor, and loss function. The SAD feature extractor is employed to capture degradation-related features from the source and target domains. The process of feature extraction $f_e$ can be formulated as follows:
\begin{equation}
	\mathbf{F}_s=f_e\left(\mathbf{X}_s\right) \quad \mathbf{F}_t=f_e\left(\mathbf{X}_t\right),
\end{equation}
where $\mathbf{F}_s$ and $\mathbf{F}_t$ are the degradation-related features.
Then, the MK-MMD is introduced to measure the distributional discrepancy between $\mathcal{P'}(\mathbf{F}_s)$ and $\mathcal{Q'}(\mathbf{F}_t)$\cite{han2022end}. By minimizing the MK-MMD, the samples in the source and target domains are transformed into a feature space where their distributions are as similar as possible. Finally, the subspace's feature $\mathbf{f}_i^s$ and $\mathbf{f}_i^t$ are further input to the CNN predictor $f_p$ to estimate the battery SOH $\hat{y}_i^s$ and $\hat{y}_i^t$, respectively, which is formulated as,
\begin{equation}
	\hat{y}_i^s = f_p\left(\mathbf{f}_i^s\right)  \quad  \hat{y}_i^t = f_p\left(\mathbf{f}_i^t\right) .
\end{equation}

\subsubsection{Self-attention distillation feature extractor}\label{feature_extractor}
\begin{figure}[!h]
	\centering
	\includegraphics[width=0.8\textwidth]{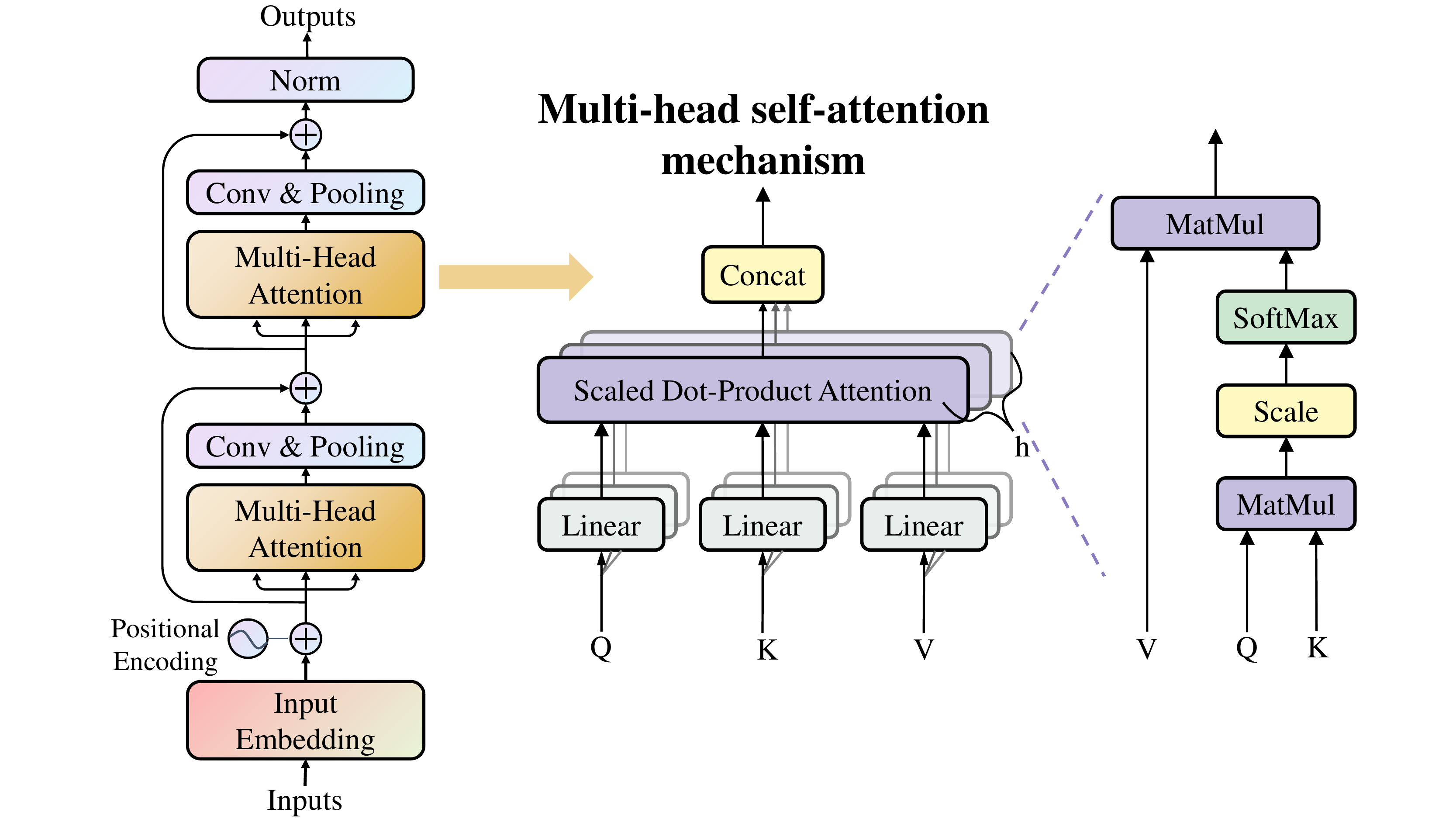}
	\caption{The structure SAD feature extractor.}
	\label{SAD}
\end{figure}

The SAD feature extractor module is designed to extract degradation-related features from the raw data automatically. The SAD module is composed of the multi-head attention mechanism and the distillation operation\cite{zhou2021informer}, as shown in Fig.~\ref{SAD}. For the battery SOH estimation task, assume that the input sequence is $ \mathbf{X} \in \mathbb{R}^{n \times m}$, where n is the length of time series and m is the number of sensors. The SAD module first adds a position encoding to the sequence to preserve the temporal information. The  fixed position encoding maps all positions into a matrix $\mathbf{P}  \in \mathbb{R}^{n \times d_{\text {model }}}$ and is defined as,
\begin{equation}
	\begin{aligned}
		\mathbf{P}_{(k, 2 j)} & =\sin \left(k /\left(2 n\right)^{2 j / d_{\text {model }}}\right) \\
		\mathbf{P}_{(k, 2 j+1)} & =\cos \left(k /\left(2n \right)^{2 j / d_{\text {model }}}\right), 
	\end{aligned}
\end{equation}
where $j \in\left\{1, \ldots,\left\lfloor d_{\text {model }} / 2\right\rfloor\right\}$, $\text {model }$ is the encoding dimension, $\mathbf{P}_{(k, 2 j)}$ and $\mathbf{P}_{(k, 2 j+1)}$ are the $2j$-th, $(2j+1)$-th components of the encoding vector at position $k$, respectively. To extract more information, a one-dimensional convolution $\operatorname{Conv1d}$ with a kernel size of 3 is applied to transform the final vector dimension of input from $m$ to $d_{\text {model }} $. Then, the input of the multi-head attention layer is 
\begin{equation}
	\mathbf{G}=\mathbf{P} + \operatorname{Conv1d}\left(\mathbf{X}\right).
\end{equation}

The self-attention mechanism excels at modeling the intrinsic relevance of time-series data. It obtains important degradation information by a weighted sum of the attention weights calculated at each position. To enhance the learning capability of the model, the self-attention mechanism is performed on the inputs projected to different subspaces, which is called the multi-head self-attention mechanism. Formally, the multi-headed self-attention mechanism projects the input $\mathbf{G}$ to the queries ($\mathbf{Q}_i$), keys ($\mathbf{K}_i$) and values ($\mathbf{V}_i$) through multiple randomly initialized weight matrices,
\begin{equation}
	\mathbf{Q}_i=\mathbf{G} \mathbf{W}_i^Q, \quad \mathbf{K}_i=\mathbf{G} \mathbf{W}_i^K, \quad \mathbf{V}_i=\mathbf{G} \mathbf{W}_i^V,
\end{equation}
where $j \in\left\{1, 2, \ldots, h \right\}$, $h$ is the number of heads, $\mathbf{W}_i^Q \in \mathbb{R}^{d_{\text {model}} \times d_q}$, $\mathbf{W}_i^K \in \mathbb{R}^{d_{\text {model}} \times d_k}$, and $\mathbf{W}_i^V \in \mathbb{R}^{d_{\text {model}} \times d_v} $ are learned in the training process and the dimension of each matrix satisfies $d_q=d_k=d_v=d_{\text {model }} / h$. 
The scaled dot-product attention is then applied to calculate the similarity between $\mathbf{Q}_i$ and $\mathbf{K}_i$ to obtain a weight matrix that measures the importance of different degradation-related features in the sequence. The weighted sum of $\mathbf{V}_i$ is the output of the $i$-th head, which is expressed as,
\begin{equation}
	\text{head}_i = \operatorname{Attention}\left(\mathbf{Q}_i, \mathbf{K}_i, \mathbf{V}_i\right)=\operatorname{softmax}\left(\frac{\mathbf{Q}_i \mathbf{K}_i^T}{\sqrt{d_k}}\right) \mathbf{V}_i \;.
\end{equation}
After that, the self-attention mechanism is calculated on the input $\mathbf{G}$ $h$ times in parallel, and all the outputs are integrated together to generate the multi-head features $\mathbf{H}^j$ at $j$-th layer,
\begin{equation}
	\mathbf{H}^j=\operatorname{MultiHead}(\mathbf{G})=\operatorname{concat}\left(\text { head }_1, \ldots, \text { head }_h\right) \;.
\end{equation}

Considering the self-attention mechanism's computational complexity and large memory consumption, the distillation operation is designed to extract dominant features in $\mathbf{H}^j$ to shrink the output length.
The distillation operation is expressed as,
\begin{equation}
	\mathbf{F}^{j}=\operatorname{MaxPool}\left(\operatorname{ELU}\left(\operatorname{Conv1d} \left(\mathbf{H}^j\right)\right)\right) \;,
\end{equation}
where $\operatorname{Conv1d}(\cdot)$ performs a one-dimensional convolution with the kernel size of 3 on temporal dimension, and $\operatorname{ELU}$ is a nonlinear activation function. The max-pooling layer with the step stride of 2 is adopted to remove redundant features and increase the domain-invariant property of the output.
% $[\cdot]_{\mathrm{AB}}$ denotes the multi-head self-attention operations

\subsubsection{Convolutional neural network predictor with smooth constraint}\label{predictor}
The predictor is applied to map the features $\mathbf{F}$ in the subspace to the battery SOH. As shown in Fig.~\ref{SKDAN_structure}, the CNN with smoothing constraint serves as a predictor in this study. Generally, the structure of CNN comprises convolutional layers and pooling layers, and the formula can be written as,
\begin{equation}
	\begin{aligned}
		& \mathbf{U}_c^l=f\left(\mathbf{W}_{c} \otimes \mathbf{F}^{l-1}+\mathbf{b}_{c}\right) \\ 
		& \mathbf{F}^l = \operatorname{MaxPool}\left(\mathbf{U}_c^l, p, t\right),
	\end{aligned}
\end{equation}
where $\otimes$ denotes the convolution operator, $f$ represents the nonlinear activation function, $\mathbf{W}_{c}$, $\mathbf{b}_{c}$ and $\mathbf{U}_c^l$ are the parameters, bias, and output of the convolutional operation, respectively. $\mathbf{F}^{l-1}$ and $\mathbf{F}^l$ are the feature map in the 
$\left(l-1\right)$-th and $l$-th layers. The pooling size $p$ and  step size $t$ are set to 4 in the max-pooling layer.
After two convolutional and pooling layers, the output of CNN is reshaped to $\mathbf{z}$ and then fed to feed-forward neural network (FNN) with a dropout technique to estimate battery SOH $y$,
\begin{equation}
	y=\operatorname{\text{ReLU}}\left(\mathbf{z} \mathbf{W}_1+\mathbf{b}_1\right) \mathbf{W}_2+\mathbf{b}_2 \;,
\end{equation}
where $\mathbf{W}_1$ and $\mathbf{W}_2$ are weight matrices, $\mathbf{b}_1$ and $\mathbf{b}_2$ are the bias, and $\operatorname{ReLU}$ represents the nonlinear activation function. 

To improve the robustness and generalization of the model, we incorporate the smoothness constraint on the predictor\cite{joseph2022lama}. The smoothness constraint comes from an intuitive concept that, if the mapping relationship is uniformly continuous, the output remains in the neighborhood of $f_p\left(\mathbf{F}\right)$ despite a small perturbation of the input $\mathbf{F}$. Accordingly, we design a smooth loss $\mathcal{L}_\text{smooth}$,
\begin{equation}
	\mathcal{L}_{\text {smooth }}=\left\|(f_p\left(\mathbf{F}\right)-f_p\left(\mathbf{F}+\gamma_{\text {noise }}\mathbf{ \delta}\right)\right\|^2 \; ,
\end{equation}
where $\mathbf{\delta} \in \mathcal{N}(0,1)$ is a is the Gaussian perturbation, and $ \gamma_{\text {noise }} $ is the scale factor that controls the size of the perturbation.

\subsubsection{Loss and optimization}\label{loss}
The optimization objective $\mathcal{L}_\text{overall}$ of the SKDAN model consists of the prediction loss $\mathcal{L}_\text{pre}$, the domain adaptation loss $\mathcal{L}_\text{MK-MMD}$ , and the smooth loss $\mathcal{L}_\text{smooth}$. The overall loss function is expressed as,
\begin{equation}\label{loss_equation}
	\mathcal{L}_\text{overall}= \mathcal{L}_\text{pre} +\lambda \mathcal{L}_\text{MK-MMD}  + \beta \mathcal{L}_\text{smooth} \;,
\end{equation}
where $\lambda$ and $\beta$ are non-negative trade-off parameters. 
Since the samples from the source domain have labels, the prediction loss $L_\text{pre}$ is calculated from the standard mean squared error (MSE) between $\hat{y}_i^s$ and $y_s^i$,
\begin{equation}
	\mathcal{L}_\text{pre}=\frac{1}{N_s} \sum_{i=1}^{N_s}\left(y_s^i-\hat{y}_s^i\right)^2.
\end{equation}

The domain adaptation loss $\mathcal{L}_\text{MK-MMD}$ is the squared distance of $\mathbf{F}_s$ and $\mathbf{F}_t$ embedded in the reproducible kernel Hilbert space (RKHS), which can be expressed as,
\begin{equation}
	\begin{aligned}
		\mathcal{L}_\text{MK-MMD} &= \left\|\mathbf{E}_{\mathcal{P}^{\prime}}\left(\phi\left(\mathbf{F}_s\right)\right)-\mathbf{E}_{\mathcal{Q}^{\prime}}\left(\phi\left(\mathbf{F}_t\right)\right)\right\|_{\mathcal{H}_k} \\
		& = \frac{1}{N_s^2} \sum_{i=1}^{N_s} \sum_{j=1}^{N_s} \phi\left(\mathbf{f}_i^s, \mathbf{f}_j^s\right)+
		\frac{1}{N_t^2} \sum_{i=1}^{N_t} \sum_{j=1}^{N_t} \phi\left(\mathbf{f}_i^t, \mathbf{f}_j^t\right) -  \\
		&  \quad \;  \frac{2}{N_s N_t} \sum_{i=1}^{N_s} \sum_{j=1}^{N_t} \phi\left(\mathbf{f}_i^s, \mathbf{f}_j^t\right)
	\end{aligned}
\end{equation}
where ${\mathcal{H}_k} $ is the RKHS with kernel $\mathcal{K}$, $\phi(\cdot)$ is a kernel-based mapping function, $\mathbf{f}_i^s$ and $\mathbf{f}_i^s$ are the subspace's feature learned from the $\mathbf{x}_i^s$ and $\mathbf{x}_i^t$, respectively. Since the single-kernel method quantifies the probability distribution of features from only one aspect, multiple kernels $k$ are weighted together to comprehensively measure the distance between different domains.
\begin{equation}
	\mathcal{K}  \triangleq\left\{k=\sum_{u=1}^m \alpha_u k_u: \sum_{u=1}^m \alpha_u=1, \alpha \geq 0, \forall u\right\}\;,
\end{equation}
where $k_u$ and $\alpha_u$ are the $u$-th kernel and the corresponding coefficients, respectively.
The network parameters are updated by optimizing the overall loss function $\mathcal{L}_\text{total}$ with the back-propagation algorithm. When the network training is completed, the network is able to estimate the SOH $\hat{y}_i^t$ of the target domain online, that is,
\begin{equation}
	\hat{y}_i^t = f_p\left( f_e\left(\mathbf{x}_i^t\right)\right).
\end{equation}

\section{Experiments and Results}\label{experiment} %% {{{
In this section, we first present two battery datasets and experimental implementations in detail. Second, we verify the effectiveness of the proposed method under different source and target domains and compare it with other state-of-the-art methods.

\subsection{Experimental setup}
\subsubsection{Data generation}
The CALCE and the SNL datasets are employed to investigate the performance of the proposed model. The batteries were tested in a high-dimensional parameter space, including temperature, DOD, discharge rate, and mean SOC over long-term cycling. The CALCE dataset consists of 8 commercial $\text{LiCoO}_2$/graphite batteries with a nominal capacity of 1.5 Ah. The experiments were conducted in the four SOC ranges of 0\%-60\%, 20\%-80\%, 40\%-100\% and 0\%-100\% with a discharge rate of C/2 and a temperature-controlled chamber at 25±2 $^{\circ}$C. The SNL dataset comprises 24 batteries lithium iron phosphate batteries from A123 Systems. The batteries are cycled at the two SOC ranges of 20\%-80\% and 0\%-100\% with various temperatures (15 $^{\circ}$C, 25 $^{\circ}$C and 35 $^{\circ}$C) and discharge rates (0.5C, 1C, 2C and 3C), as referred to in Table~\ref{battery_list}. 
\begin{table*}[!htb]
	\rmfamily
	\caption{Cycled batteries and cycling conditions for the data generation}
	\newcolumntype{L}[1]{>{\raggedright\arraybackslash}p{#1}}
	\resizebox{\textwidth}{!}{
		\begin{tabular}{L{1.5cm}L{1.5cm}L{2cm}L{1.5cm}L{2cm}L{2cm}L{3cm}L{2cm}}
			\hline
			\specialrule{0em}{1pt}{1pt}  
			Datasets & Materials & Nominal capacity  & Voltage range & SOC range & cycling temperature &  Charge/discharge current rate (C) &  Number of batteries   \\
			\hline
			\specialrule{0em}{1pt}{1pt}  
			\multirow{4}{*}{CALCE} & \multirow{4}{*}{$\operatorname{LiCoO_2}$} &  \multirow{4}{*}{1.5Ah}  & \multirow{4}{*}{2.75-4.2V} & 0\%-100\% & \multirow{4}{*}{25$^{\circ}$C} & \multirow{4}{*}{0.5/0.5} & 2\\
			& & & &  0\%-60\%    & & & 2 \\
			& & & &  20\%-80\%    & & & 2 \\
			& & & &  40\%-100\%    & & & 2 \\
			\multirow{8}{*}{SNL} & \multirow{8}{*}{$\operatorname{LiFePO_4}$} &  \multirow{8}{*}{1.1Ah}  & \multirow{8}{*}{2-3.6V} & \multirow{7}{*}{0\%-100\%} & 15$^{\circ}$C & 0.5/1 & 2\\
			& & & &  & 15$^{\circ}$C & 0.5/2  & 2 \\
			& & & &  & 25$^{\circ}$C & 0.5/1  & 4 \\
			& & & &  & 25$^{\circ}$C & 0.5/2  & 2 \\
			& & & &  & 25$^{\circ}$C & 0.5/3  & 4 \\
			& & & &  & 35$^{\circ}$C & 0.5/1  & 4 \\
			& & & &  & 35$^{\circ}$C & 0.5/2  & 2 \\
			& & & & 20\%-80\% & 25$^{\circ}$C & 0.5/0.5  & 4 \\
			\hline
			\specialrule{0em}{1pt}{1pt}  
	\end{tabular}}
	\label{battery_list}
\end{table*}

\begin{figure}[!h]
	\centering
	\subfigure[Voltage and current profiles in the characterization test]{\label{protocol}
		\includegraphics[width=0.67\textwidth]{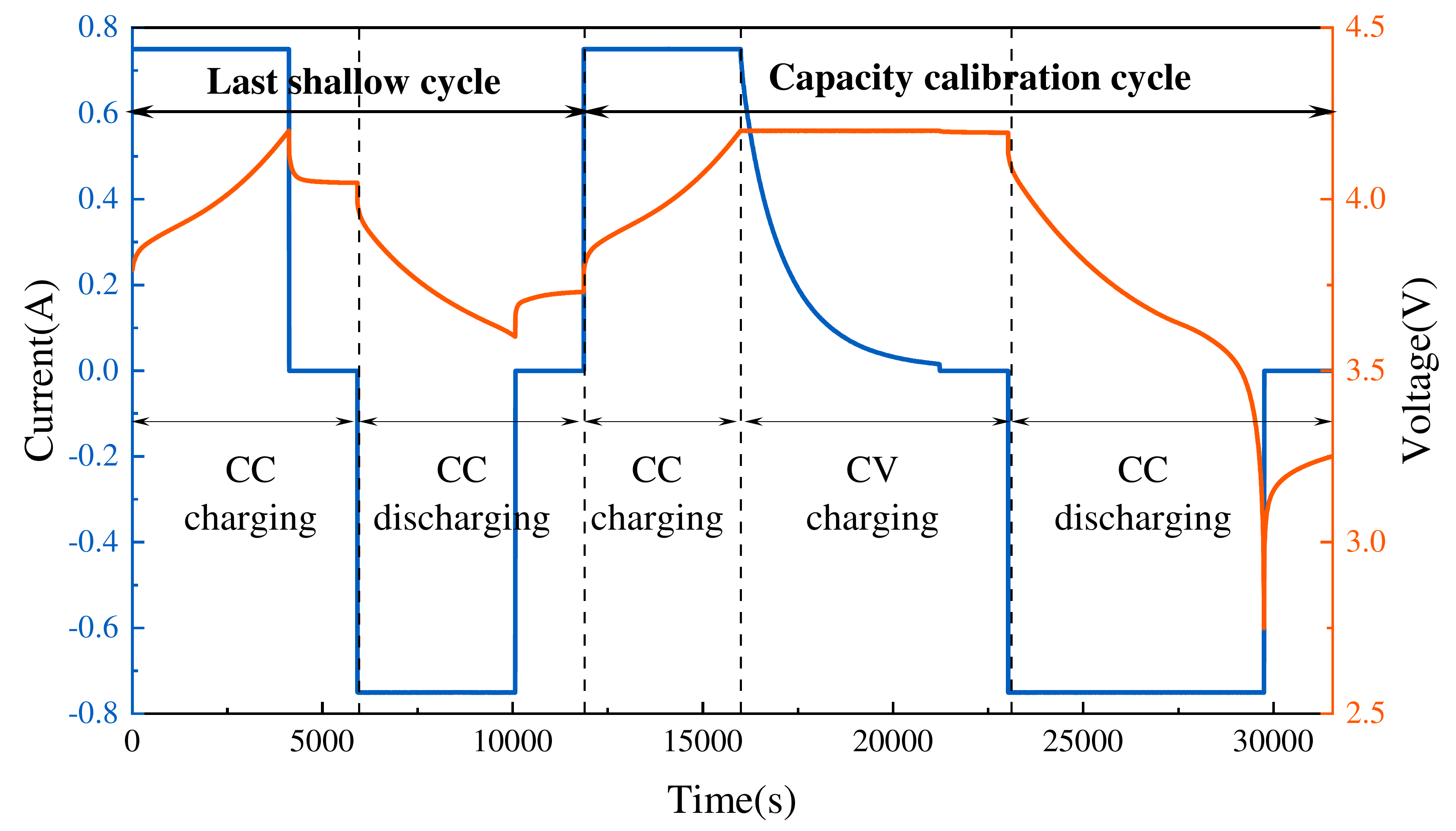}}
	\subfigure[SNL battery degradation curves]{\label{SNL_degradation}
		\includegraphics[width=0.3\textwidth]{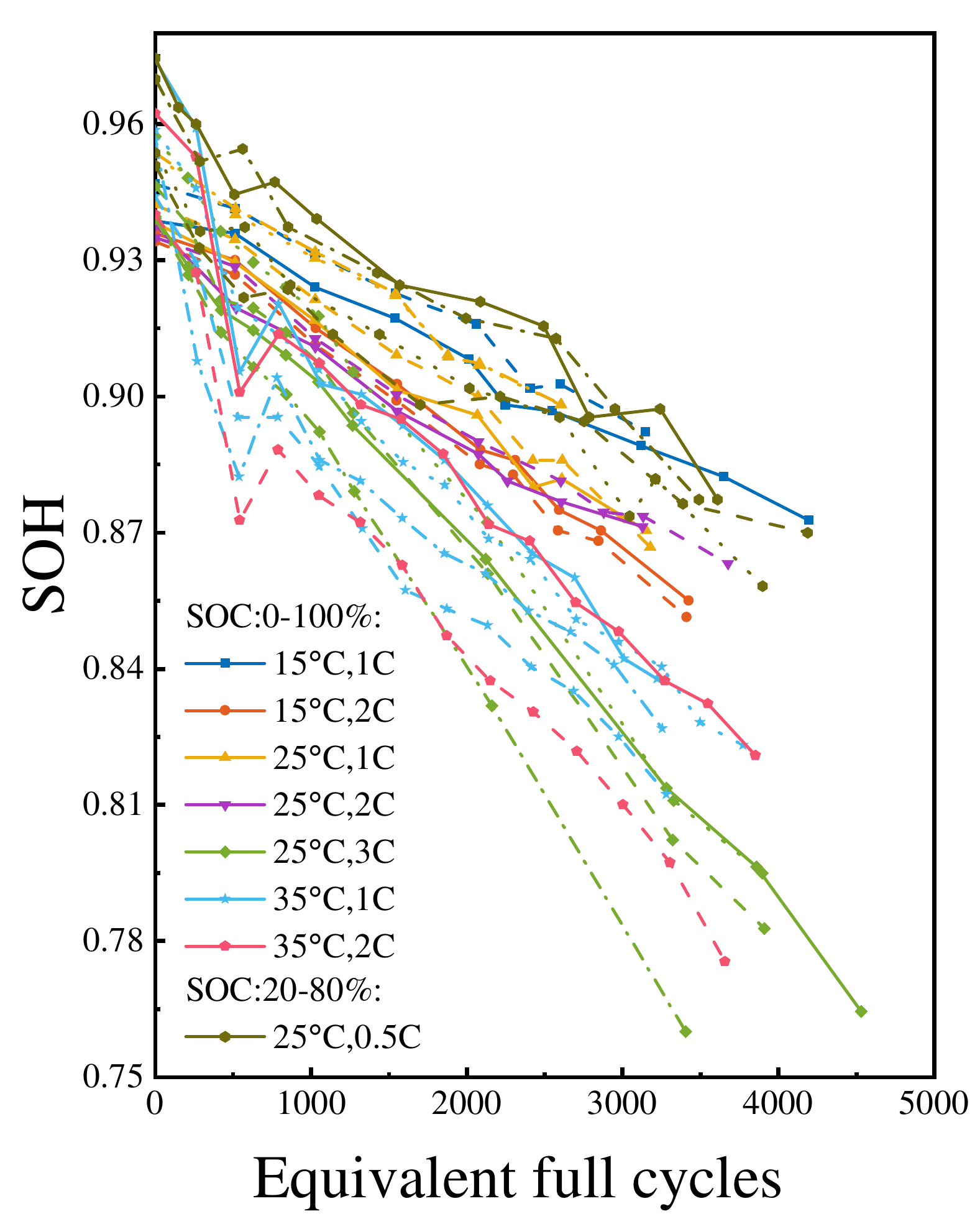}}
	\caption{Battery cycling data}
\end{figure}

For the batteries operating under shallow-cycle conditions, a characterization test is required after every 50/100 shallow cycles to determine the present maximum capacity of the battery, referred to as calibrated capacity. The characterization test involves a specific full cycle, which is a 0.5C constant current (CC) discharge to cut-off voltage after a 0.5C constant current constant voltage (CC-CV) protocol charges to 100\% SOC, as shown in Fig.\ref{protocol}. According to the results of characterization test, the battery capacity decay curves of the CALCE and the SNL dataset with different SOC ranges and operating conditions are shown in Fig.~\ref{direct_estimation} (a) and Fig.~\ref{SNL_degradation}. Finally, the calibrated capacity represents the true capacity of the battery in the last shallow cycle and is applied to evaluate the accuracy of the model estimation results. A more detailed description of charge/discharge protocols and analysis for each battery dataset can be found in\cite{saxena2016cycle,preger2020degradation}.

\subsubsection{Data preprocessing}
We select the CC phase of the charge process as the input samples to estimate battery SOH.  Since the length of each cycle varies for different SOC ranges, the moving window is first introduced to split the full cycles into segments. Specifically, the size of the sliding window is set to the DOD of the battery at shallow-cycle conditions, and the step size is 10\% SOC range. To ensure consistency across different segments, all segments are further resampled to 160 points as each segment contains a different charge duration. The full cycles are eventually divided into segments that are as long as the shallow cycles.

To accurately describe the charge segment, the four sequences: voltage ($\mathbf{v}$), voltage difference ($\Delta \mathbf{v}$), capacity difference ($\Delta \mathbf{q}$), and incremental capacity (IC), are adopted in this paper. Given any part of the charge segment, the voltage $\mathbf{v}=\left[v_1, v_2, \ldots, v_n\right]$ and current $\mathbf{i}=\left[i_1, i_2, \ldots, i_n\right]$ can be directly measured by the BMS with a fixed sampling time $\Delta t$. The voltage difference is derived from the voltage $\mathbf{v}$ and can be expressed as $\Delta \mathbf{v}=\mathbf{v} - v_1$. The battery capacity $\mathbf{q}$ is obtained by integrating the current with time. However, as the capacity is unknown at the beginning of the battery charge process, we can only obtain the capacity difference $\Delta \mathbf{q}$ based on available data. Suppose that $\Delta q_j$ represents the $j$-th component of the $\Delta \mathbf{q}$ sequence, and its formula is as follows:
\begin{equation}
	\Delta q_j=\int_{t=t_1}^{t=t_j}I dt,
\end{equation}
where $t_1$ and $t_j$ are the starting and current times, respectively.
The IC curve is generated by differential processing of the capacity-voltage curve, and is defined as,
\begin{equation}
	IC_j=\left\{
	\begin{array}{lcl}
		\frac{d q_j}{d v_j} \approx \frac{\Delta q_j-\Delta q_{j-1}}{\Delta v_j-\Delta v_{j-1}} & &{j=2,\cdots,n}\\
		0 & & {j=1}
	\end{array} \right.
\end{equation}
The IC curve transforms voltage plateau areas into easy-to-identify peaks, effectively reflecting the evolution of battery degradation. Accordingly, the input sample $\mathbf{X}$ is represented as $\mathbf{X}=\left[\mathbf{v};\Delta \mathbf{v}; \Delta \mathbf{q}; \operatorname{IC}\right]$. Regardless of how the operating range of the battery SOC is set, the $\mathbf{X}$ can always be extract from the charge process and fed to the model to estimate battery SOH. Furthermore, to eliminate the effects of magnitude, the data in the source and target domains are individually normalized to the $\left[0, 1\right]$ range using min-max normalization:
\begin{equation}
	\tilde{\mathbf{x}}_j=\frac{\mathbf{x}_j-\min \left(\mathbf{x}_j\right)}{\max \left(\mathbf{x}_j\right)-\min \left(\mathbf{x}_j\right)},
\end{equation}
where $\tilde{\mathbf{x}}_j$ and $\mathbf{x}_j$ denotes the normalized and original vector of all inputs of the $j$-th feature, respectively.

\subsubsection{Training procedure}
The performance of the SKDAN model is evaluated using the CALCE and SNL datasets, where 50\% of the batteries are used for training and 50\% for testing. The SKDAN model minimizes the loss defined in Eq.~\ref{loss_equation} in the training set by applying back-propagation algorithm and Adam optimizer to update the network weights. The maximum number of training epochs is preset to 200, and other network hyperparameters, such as learning rate, batch size, and trade-off rate, are optimized through a random search algorithm. For each transfer experiment, the random search algorithm automatically generates 100 possible configurations in a preset hyperparameter space and measures their performance. Taking the SKDAN model trained in CALCE data as an example, the optimal hyperparameters settings are presented in Table \ref{hyperparameter}.

\begin{table*}[!htb]
	\rmfamily
	\caption{Hyperparameter configurations of SKDAN for the CALCE dataset}
	\newcolumntype{L}[1]{>{\raggedright\arraybackslash}p{#1}}
	\resizebox{\textwidth}{!}{	
	\begin{tabular}{L{5.5cm}L{4cm}L{4cm}L{4cm}}
		\hline
		\specialrule{0em}{1pt}{1pt}  
		Hyperparameter & 0\%-60\% & 20\%-80\%  & 40\%-100\% \\
		\hline
		\specialrule{0em}{1pt}{1pt}  
		Batch size & 16 & 64 & 32  \\
		Learning rate &  $7.5\times10^{-4}$  & $5.6\times10^{-5}$ &  $4.7\times10^{-3}$ \\
		Number of attention layers & 3 & 2 & 2  \\
		Size of attention dimension & 128 & 128 & 128 \\
		Number of attention heads & 4 & 2 & 2 \\
		Kernel size & 2 & 3 &  5 \\
		Number os neurons in FNN & 32 & 64 & 16 \\
		Dropout rate&  0.2 & 0.3 & 0.4 \\
		Smoothness weight& 0.08 & 0.05 & 0.12 \\
		MK-MMD weight& 0.72 & 1.33 & 1.06 \\
		\hline
		\specialrule{0em}{1pt}{1pt}  
	\end{tabular}}
	\label{hyperparameter}
\end{table*}

\subsubsection{Evaluation metrics}
To comprehensively evaluate the accuracy of the SKDAN model, three evaluation metrics, root-mean-square error (RMSE), mean absolute error (MAE), and score, are chosen to quantify the estimation error. The MAE and RMSE are commonly used performance metrics in battery SOH estimation, and are defined as follows,
\begin{equation}
	\mathrm{RMSE} =\sqrt{\frac{1}{n} \sum_{i=1}^n\left(\hat{y}_i-y_i\right)^2}\;,
\end{equation}
\begin{equation}
	\mathrm{MAE} =\frac{1}{n} \sum_{i=1}^n\left|\hat{y}_i-y_i\right| \;,
\end{equation}
where $n$ is the number of samples, $\hat{y}_i$ and $y_i$ are the estimated and true value of $i$-th sample. In addition, the score function in the prognostics and health management fields is introduced to evaluate the effect of overestimation ($\hat{y}_i\geq y_i$) and underestimation ($\hat{y}_i\leq y_i$)\cite{zhang2022prediction}. Since the overestimation is more likely to lead to serious accidents, while the underestimation is able to provide a margin of safety, the score function penalizes errors from overestimation more than errors from underestimation. The score of the test dataset containing $n$ samples is expressed as follows:
\begin{equation}
	s= \begin{cases}\sum\limits_{i=1}^n\left(e^{-\frac{d_i}{1.3}}-1\right), & d_i<0 \\ \sum\limits_{i=1}^n\left(e^{d_i}-1\right), & d_i \geq 0\end{cases},
\end{equation}
where $d_i = \hat{y}_i - y_i$ represents the difference between the $i$-th sample's estimated value and true value.

\subsection{Performance evaluation}
In this subsection, we conducted a series of experiments that transfer knowledge from full cycles to shallow cycles with different SOC ranges in the CALCE dataset, from full cycles with different working conditions to shallow cycles in the SNL dataset, and transfer across the CALCE and SNL datasets. All experiments were written in python 3.8 with Pytorch 1.9.0 deep learning toolkit and performed on a high-performance computing platform with Intel(R) Xeon(R) E5-2620 v3 CPU. 

\subsubsection{Single domain knowledge transfer for different SOC ranges}
\begin{figure}[!h]
	\centering
	\subfigure[0-60\% SOC range]{
		\includegraphics[width=0.32\textwidth]{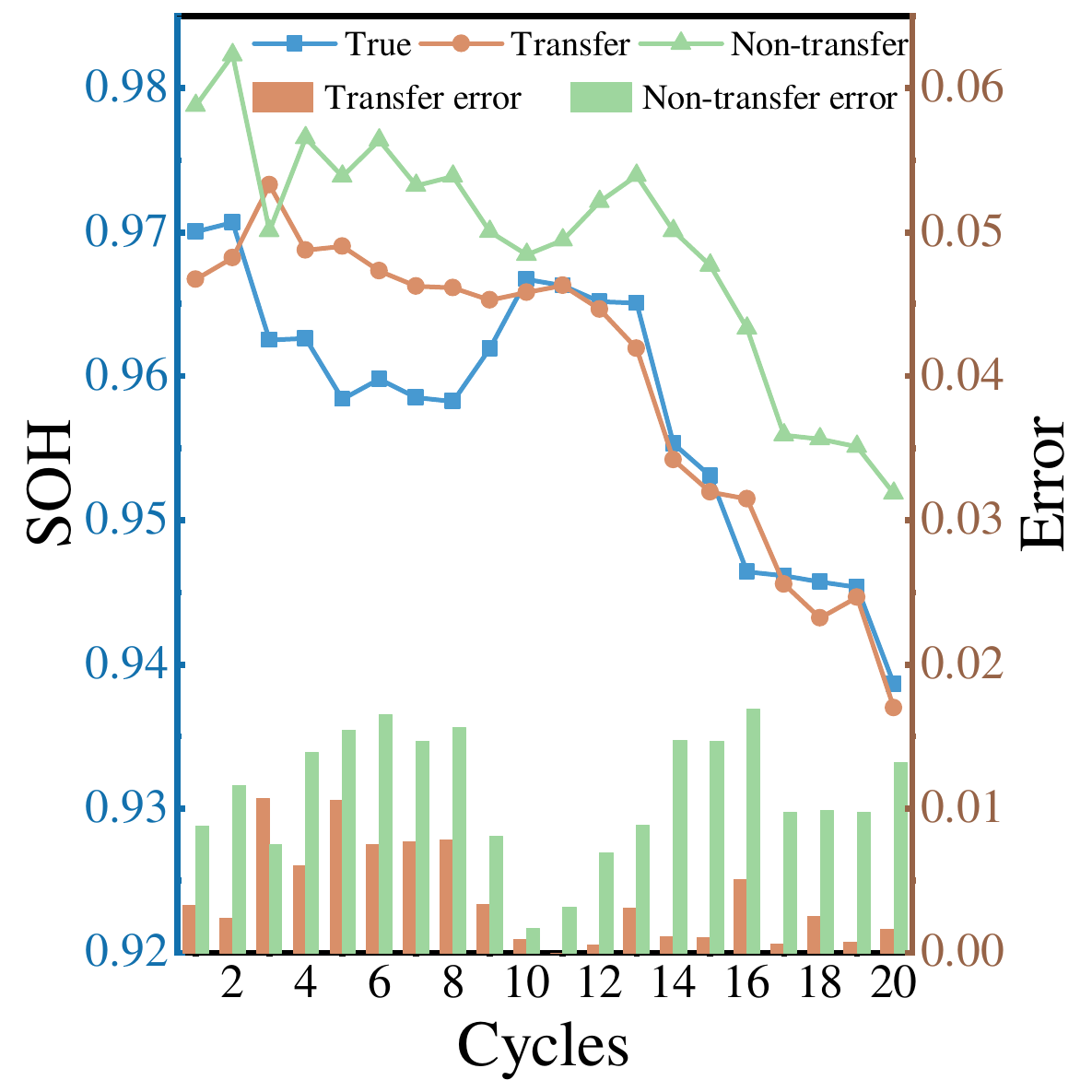}} 
	\subfigure[20-80\% SOC range]{
	\includegraphics[width=0.32\textwidth]{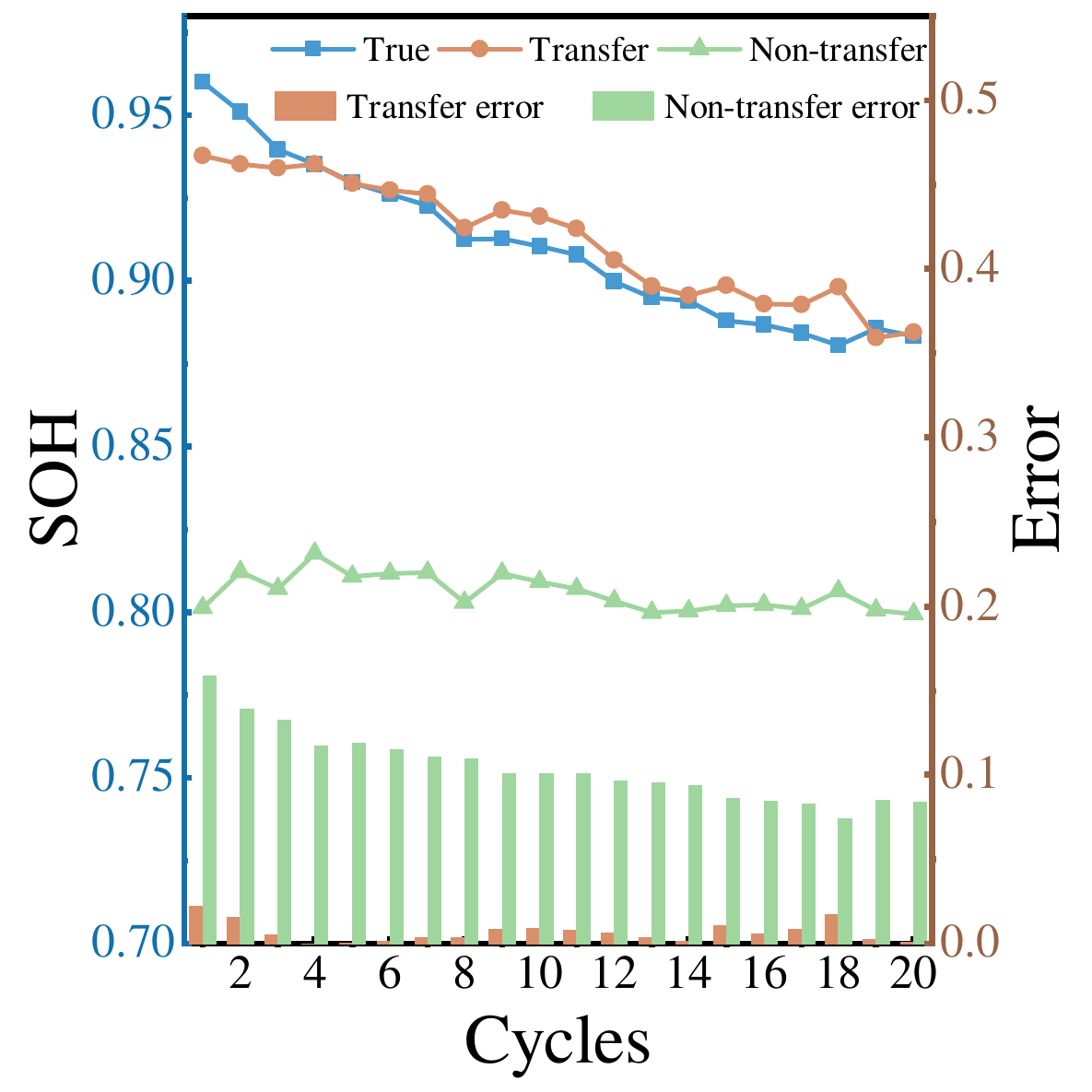}}
	\subfigure[40-100\% SOC range]{
	\includegraphics[width=0.32\textwidth]{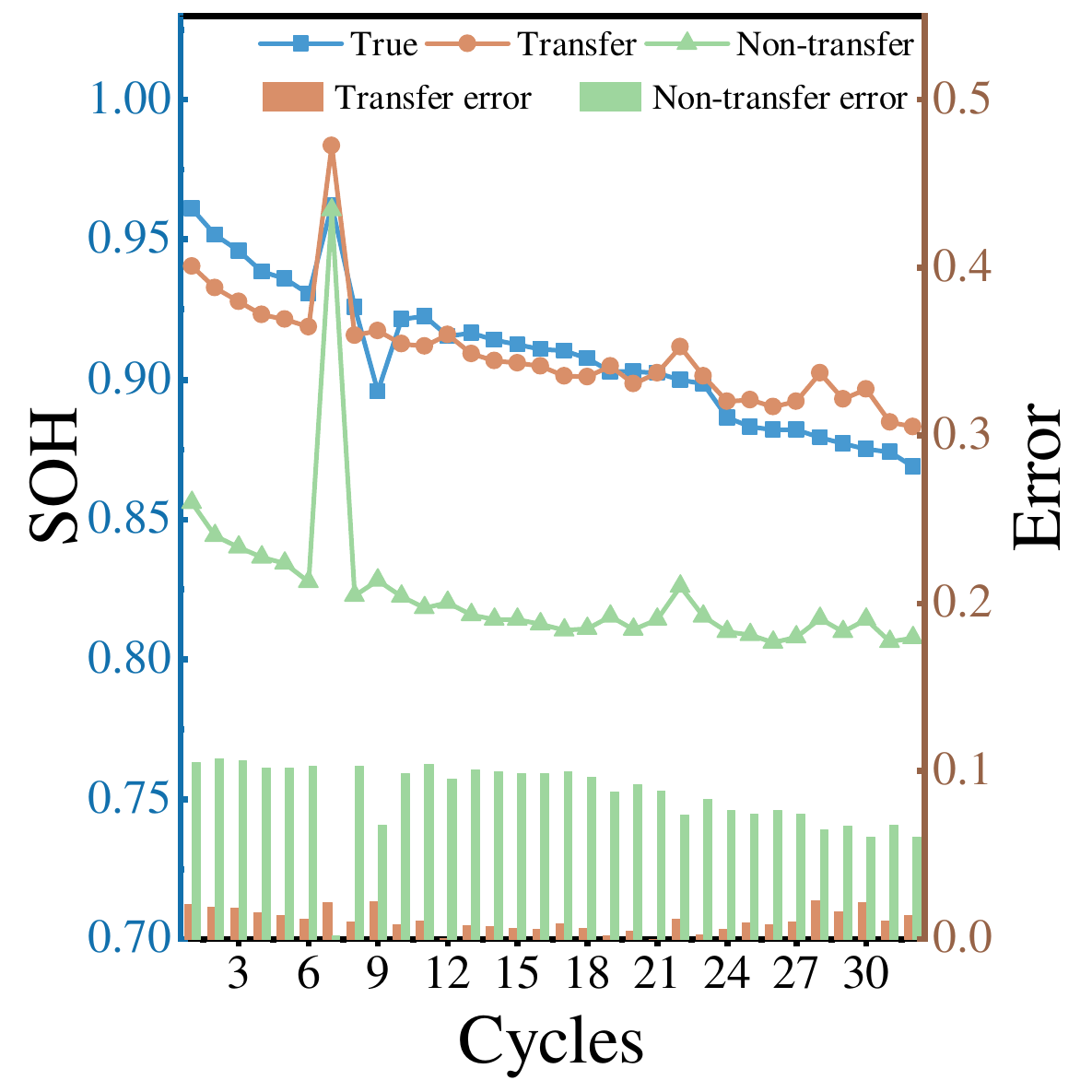}}
	\caption{SOH estimation results on the CALCE dataset}
	\label{CALCE_transfer} %% label for entire figure
\end{figure}

The CALCE dataset is employed to investigate the effectiveness of the proposed SKDAN model for capacity estimation under shallow cycles with different SOC ranges. The experiments take 0.5C, 0\%-100\% full-cycles data as the source domain, and 0\%-60\%, 20\%-80\%, and 40\%-100\% shallow-cycles data as the target domain, respectively. The SKDAN model is first compared with a non-adaptation SKDAN (Nonad-SKDAN) model which is trained with only source domain data. To enhance the model's reliability, each transfer experiment pair is run ten times and takes the average value as the estimation result. As shown in Fig.~\ref{CALCE_transfer}, the SKDAN model is able to track the true SOH more closely than the non-adaptation SKDAN model. For the shallow-cycle batteries in the 0\%-60\% SOC range, since the SOH of batteries remains at a high level (SOH > 98\%) and the domain discrepancy with full-cycle batteries is small, the SKDAN model only shows a slight improvement in performance. For two transfer experiments of 20\%-80\% and 40\%-100\%, the SKDAN model exhibits a significant improvement in accuracy, reducing RMSE (-10.76\%, -12.89\%), MAE (-10.34\%, -12.41\%), and score (-2.64, -1.95). This indicates the SKDAN model can effectively learn domain-invariant features to overcome the challenge of domain discrepancy.

The SKDAN model is further compared with the other seven models to demonstrate its superiority in performance. The other models are UDTL-based architectures but with different feature extractors, predictors and domain adaptation methods. They are CNN-BiLSTM\cite{shen2023source}, BiLSTM-CNN, SA-FNN, SA-LSTM, and SA-BiLSTM\cite{zhang2022predictionbi} with MK-MMD, SKDAN with MMD, and SKDAN with correlation alignment (CORAL)\cite{su2022hybrid}. For a fair comparison, the random search algorithm is adopted for each model to select the optimal hyperparameters. The mean and standard deviations of the RMSE, MAE, and score metrics for all models are listed in Table \ref{resultcompare}.

\begin{table*}[!htb]
	\rmfamily
	\caption{Results of the proposed SKDAN and other state-of-the-art methods on the CALCE dataset}
	\newcolumntype{L}[1]{>{\raggedright\arraybackslash}p{#1}}
	\resizebox{\textwidth}{!}{
		\begin{tabular}{L{1.4cm}L{1.4cm}L{1cm}L{1.7cm}L{1.7cm}L{1.7cm}L{1.7cm}L{1.7cm}L{1.7cm}L{1.7cm}L{1.7cm}}
			\hline
			\specialrule{0em}{1pt}{1pt}  
			Source & Target & Metrics & CNN-BiLSTM &  BiLSTM-GRU &  SA-FNN & SA-LSTM & Nonad-SKDAN & SKDAN-CORAL  & SKDAN-MMD & \textbf{Ours}\\
			
			\hline
			\specialrule{0em}{1pt}{1pt}  
			\multirow{3}{*}{0\%-100\%} &\multirow {3}{*}{0\%-60\%} & RMSE & $1.65 \pm 0.50$ & $1.53 \pm 0.45$& $1.37 \pm 0.24$ & $1.35 \pm 0.41$ & $1.75 \pm 0.55 $ &$  1.31 \pm 0.38$&   $  1.00 \pm 0.31    $ & $   1.01 \pm 0.19$    \\
			&  & MAE  &$ 1.53 \pm 0.45   $&$ 1.30 \pm 0.43 $&$ 0.99 \pm 0.24  $&$ 1.13 \pm 0.37  $&$ 1.62 \pm 0.56  $&$ 1.05 \pm 0.31  $&$ 0.82 \pm 0.26   $&$ 0.83 \pm 0.18$\\
			&  & Score &$ 0.25 \pm 0.06   $&$  0.22 \pm 0.05  $&$ 0.20 \pm 0.05   $&$ 0.20 \pm 0.05  $&$ 0.32 \pm 0.12   $&$ 0.19 \pm 0.07   $&$  0.15 \pm 0.06   $&$ 0.15 \pm 0.04$\\
			
			\multirow{3}{*}{0\%-100\%} &\multirow {3}{*}{20\%-80\%} & RMSE & $3.84 \pm 0.28  $&$ 4.17 \pm 0.18   $&$ 2.84 \pm 0.31  $&$ 2.96 \pm 0.38  $&$ 12.61 \pm 1.48  $&$  2.96 \pm 0.77  $&$  2.44 \pm 0.46  $&$   1.85 \pm 0.18$ \\
			&  & MAE & $3.16 \pm 0.26  $&$ 3.44 \pm 0.17  $&$ 2.13 \pm 0.13  $&$ 2.48 \pm 0.30 $&$ 11.85 \pm 1.26  $&$ 2.47 \pm 0.69  $&$  2.02 \pm 0.38   $&$ 1.51 \pm 0.13$\\
			&  & Score &  $0.87 \pm 0.07  $&$ 0.94 \pm 0.03  $&$ 0.60 \pm 0.05  $&$ 0.68 \pm 0.08  $&$ 3.07 \pm 0.36   $&$ 0.68 \pm 0.16   $&$  0.57 \pm 0.11  $&$ 0.43 \pm 0.03$\\
			
			\multirow{3}{*}{0\%-100\%} &\multirow {3}{*}{40\%-100\%} & RMSE & $3.61 \pm 0.16  $&$ 3.43 \pm 0.17  $&$ 2.91 \pm 0.28  $&$ 2.63 \pm 0.38  $&$ 14.80 \pm 2.37  $&$ 2.77 \pm 0.57   $&$ 2.17 \pm 0.42   $&$ 1.91 \pm 0.19$ \\
			&  & MAE & $3.07 \pm 0.12  $&$ 2.93 \pm 0.13  $&$ 2.49 \pm 0.26 $&$  2.25 \pm 0.32  $&$ 14.07 \pm 2.24 $&$  2.34 \pm 0.51  $&$ 1.85 \pm 0.38   $&$ 1.66 \pm 0.15$\\
			&  & Score &  $0.59 \pm 0.03  $&$ 0.56 \pm 0.03  $&$ 0.46 \pm 0.06 $&$ 0.43 \pm 0.07  $&$ 2.29 \pm 0.41  $&$ 0.42 \pm 0.09  $&$  0.36 \pm 0.05   $&$ 0.34 \pm 0.04$\\
			
			\hline
			\specialrule{0em}{1pt}{1pt}  
	\end{tabular}}
	\begin{tablenotes}[flushleft]
		\footnotesize
		\item \textbf{\textit{Note}}:The RMSE and MAE in the table are multiplied by 100. The format is: mean $\pm$ std.
	\end{tablenotes}
	\label{resultcompare}
\end{table*}

It can be observed that the SKDAN model achieves better estimation results in single domain knowledge transfer. All UDTL-based models outperform the non-adaptation SKDAN model, which verifies the feasibility of transfer learning to solve battery capacity estimation under shallow cycles. It can be also seen from Table \ref{resultcompare} that the MK-MMD is more capable of aligning feature distribution in subspace than the MMD and the CORAL,  and implements smaller estimation errors. Additionally, the proposed model has an improvement in the 20\%-80\% and 40\%-100\% SOC ranges compared to the SA-FNN and SA-LSTM model, where RMSE decreases by 0.99\% and 1.11\%, MAE by 0.62\% and 0.97\%, and score by 0.17 and 0.09. In the situation of large domain discrepancy, the CNN provides a better nonlinear mapping of degradation-related features to battery SOH than LSTM and FNN.

\subsubsection{Multiple domain knowledge transfer with different operating conditions}
\begin{figure}[!h]
	\centering
	\includegraphics[width=1\textwidth]{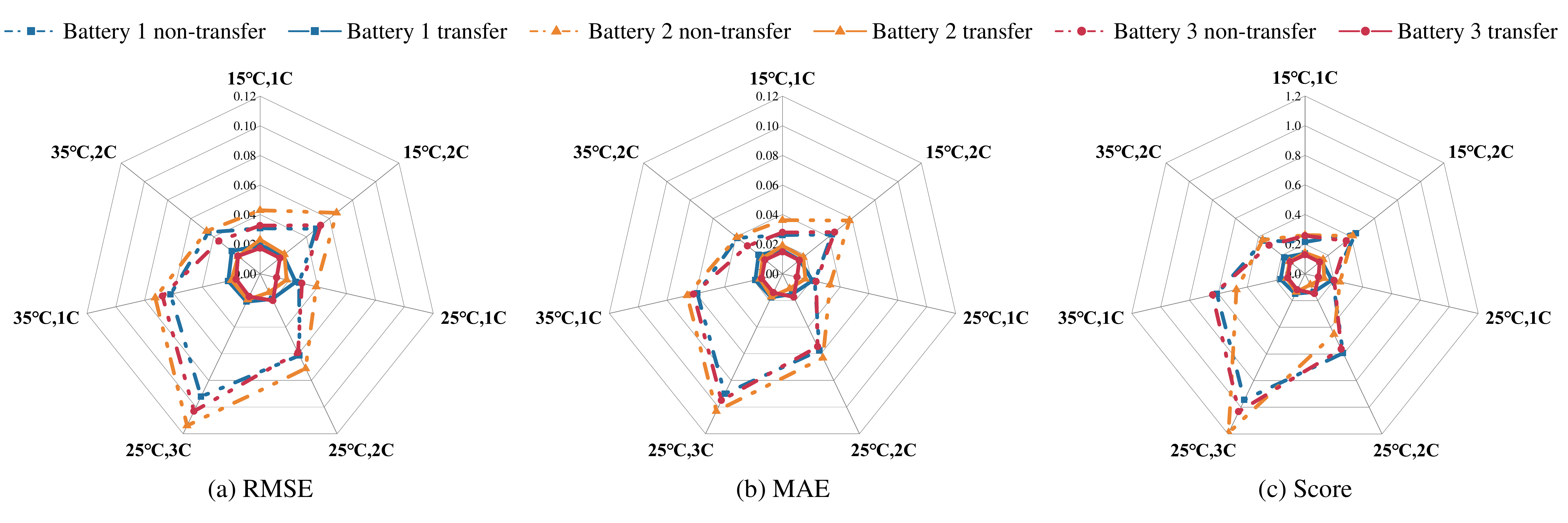}
	\caption{Radar charts of performance metrics for different operating conditions.}
	\label{multiple_domain}
\end{figure}

Battery operating conditions have a significant impact on the battery degradation profiles. The SNL dataset is applied to verify the effectiveness of the SKDAN model for transferring knowledge from full cycles with different operating conditions to shallow cycles. Here, the full cycles for each of the seven operating conditions are used as the source domain, and the three batteries under shallow cycles, with the SOC range of 20\%-80\%, are selected as the target domain. Figs.~\ref{multiple_domain} (a)-(c) shows the estimation results of the SKDAN model with and without domain adaptation on the shallow-cycle batteries.
Due to different degrees of domain discrepancy in the data at various operating conditions, the non-adaptation SKDAN model yields large estimation errors in most of  multiple domain knowledge transfer. After adding domain adaptation, the SKDAN model can effectively extract underlying consistent degradation characteristics of the battery at different conditions, so as to alleviate the effects of inconsistent data distribution caused by differences in discharge rates and temperatures. Hence, the estimation performance of the battery SOH is significantly improved under all operating conditions, verifying the robustness of the proposed method.

\subsubsection{Cross-domain knowledge transfer}
In real-world applications, collecting complete battery degradation data is a laborious task. This is especially true for brand-new batteries, where such testing is more expensive and time-consuming due to the high dimensional parameter space and long test cycles\cite{jiang2022fast,jiang2021bayesian}. These factors significantly reduce the developing speed of the data-driven model. Hence, we expect that one batch of battery degradation data can be applied to facilitate the development of SOH estimation models for batteries with different nominal capacities and electrode chemistries. Accordingly, the cross-domain transfer learning capability of SKDAN is explored on the CALCE and the SNL datasets. The experiments are conducted to transfer knowledge from the full cycles in one dataset to the shallow cycles in the range of 20\%-80\% SOC in the other dataset. The results of RMSE, MAE, and score for the eight experiments are listed in Table~\ref{cross_transfer}.
It can be seen from Table~\ref{cross_transfer} that the performance of the SKDAN model is significantly better than the SKDAN model without domain adaptation. The SKDAN model successfully captures degradation-related domain-invariant features and enables knowledge transfer of full cycles to shallow cycles between different batteries. It indicates that the SKDAN model has strong domain adaptive capability when processing the cross-domain transfer problems of batteries.

\begin{table*}[!htb]
	\rmfamily
	\caption{Cross-domain estimation results of SKDAN model with and without domain adaptation}
	\newcolumntype{L}[1]{>{\raggedright\arraybackslash}p{#1}}
	\resizebox{\textwidth}{!}{
		\begin{tabular}{L{2.2cm}L{3cm}L{1.8cm}L{1.8cm}L{1.8cm}L{1.8cm}L{1.8cm}L{1.8cm}}
			\hline
			\specialrule{0em}{1pt}{1pt}  
			\multirow{2}{*}{Source} & \multirow{2}{*}{Target} & \multicolumn{3}{c}{Nonad-SKDAN} & \multicolumn{3}{c}{SKDAN} \\
			& & RMSE & MAE & Score & RMSE & MAE & Score  \\
			\hline
			\specialrule{0em}{1pt}{1pt}  
			15$^{\circ}$C/1C (S) & 20\%-80\% SOC (C)  & $7.26\pm0.31$ & $5.91\pm0.28$ & $1.51\pm0.19$ & $2.59 \pm0.31$ & $2.07\pm0.19$ & $0.60\pm0.05$\\
			15$^{\circ}$C/2C (S) & 20\%-80\% SOC (C)  & $8.17 \pm 0.43 $ & $ 6.71 \pm 0.37$& $1.72 \pm 0.21$ & $2.61 \pm 0.17$ & $2.10 \pm 0.13$& $0.61 \pm 0.03$\\
			25$^{\circ}$C/1C (S) & 20\%-80\% SOC (C)  & $7.52 \pm 0.37$ & $6.13 \pm 0.33$ & $1.57 \pm 0.20$ & $2.57 \pm  0.15$ &  $1.90 \pm 0.13$ & $0.54 \pm 0.04$\\
			25$^{\circ}$C/2C (S) & 20\%-80\% SOC (C)  & $7.14 \pm 0.34$ & $5.84 \pm 0.31$ & $1.49 \pm 0.15$ & $2.71 \pm 0.14$ & $2.23 \pm 0.10$ & $0.65 \pm 0.03$\\
			25$^{\circ}$C/3C (S) & 20\%-80\% SOC (C)  & $8.83 \pm 0.41$ &  $7.52 \pm 0.38$ & $1.93 \pm 0.25$ & $1.91 \pm 0.25$ & $1.43 \pm 0.14$ & $0.40 \pm 0.04$ \\
			35$^{\circ}$C/1C (S) & 20\%-80\% SOC (C)  & $8.22 \pm 0.38$ & $6.99 \pm 0.34$ & $1.79 \pm 0.18$ & $2.08 \pm 0.13$ & $1.67 \pm 0.13$ &  $0.43 \pm 0.04$ \\
			35$^{\circ}$C/2C (S) & 20\%-80\% SOC (C)  & $9.42 \pm 0.47$  &  $8.14 \pm 0.42$ &  $2.09 \pm 0.23$ & $2.41 \pm 0.22$ & $1.57 \pm 0.16$ & $0.44 \pm 0.05$\\
			25$^{\circ}$C/0.5C (C) & 20\%-80\% SOC (S)  & $7.36 \pm 0.62$ & $7.12 \pm 0.56$ & $0.74 \pm 0.05$ &  $2.39 \pm 0.13$  &   $2.02\pm0.12$   &   $0.18\pm0.06$  \\

			\hline
			\specialrule{0em}{1pt}{1pt}  
  	\end{tabular}}
  	\begin{tablenotes}[flushleft]
  	\footnotesize
  	\item \textbf{\textit{Note}}:The RMSE and MAE in the table are multiplied by 100. "S" and "C" represent the SNL dataset and the CALCE dataset, respectively.
  	\end{tablenotes}
	\label{cross_transfer}
\end{table*}

\section{Discussion}\label{discussion}
\subsection{Ablation study}
The SKDAN model has shown superior performance in the battery SOH estimation of shallow-cycle batteries. To evaluate the contribution of each part of the model, we perform ablation experiments on the SKDAN model. The investigation includes four variants of the proposed model with the following differences: Model-1 omits the multi-head self-attention mechanism in the feature extractor; Model-2 has no distillation operation in the feature extractor; Model-3 replaces the CNN with FNN in the predictor; and Model-4 removes the smoothing constraint from the predictor. Each model is tested ten times on the 20\%-80\% and 40\%-100\% SOC ranges in the CALCE dataset to guarantee a fair comparison, and the mean and deviation of the results are listed in Table ~\ref{ablation_study}.

\begin{table*}[!htb]
	\rmfamily
	\caption{ Results of ablation study on the CALCE dataset}
	\newcolumntype{L}[1]{>{\raggedright\arraybackslash}p{#1}}
	\resizebox{\textwidth}{!}{
		\begin{tabular}{L{2cm}L{2cm}L{2cm}L{2cm}L{2cm}L{2cm}L{2cm}L{2cm}}
			\hline
			\specialrule{0em}{1pt}{1pt}  
			Source & Target & Metrics & Model-1 &  Model-2 &  Model-3 & Model-4 & \textbf{SKDAN}  \\
			
			\hline
			\specialrule{0em}{1pt}{1pt}  
			\multirow{3}{*}{0\%-100\%} &\multirow {3}{*}{20\%-80\%} & RMSE &  $3.67 \pm 0.36  $&$ 2.27 \pm 0.17  $&$  2.84 \pm 0.31   $&$ 2.04 \pm0.14   $&$ 1.85 \pm 0.18$ \\
			&  & MAE &  $2.94 \pm 0.20  $&$ 1.83 \pm 0.13   $&$ 2.13 \pm 0.13  $&$ 1.64 \pm 0.10 $&$ 1.51 \pm 0.13$\\
			&  & Score &  $0.82 \pm 0.06  $&$  0.51 \pm 0.03  $&$ 0.60 \pm 0.05  $&$ 0.47 \pm 0.03 $&$ 0.43 \pm 0.03$\\
			\multirow{3}{*}{0\%-100\%} &\multirow {3}{*}{40\%-100\%} & RMSE &  $3.53 \pm 0.50  $&$ 2.20 \pm 0.38 $&$ 2.91 \pm 0.28  $&$ 2.16 \pm 0.31   $&$ 1.91 \pm 0.19$ \\
			&  & MAE &  $2.90 \pm 0.41  $&$ 1.79 \pm 0.29 $&$  2.49 \pm 0.26   $&$ 1.79 \pm 0.26    $&$ 1.66 \pm 0.15$\\
			&  & Score &  $0.81 \pm 0.10  $&$ 0.52 \pm 0.10 $&$ 0.46  \pm 0.06  $&$ 0.52 \pm 0.09   $&$ 0.34 \pm 0.04$\\
			
			\hline
			\specialrule{0em}{1pt}{1pt}  
	\end{tabular}}
	\begin{tablenotes}[flushleft]
		\footnotesize
		\item \textbf{\textit{Note}}:The RMSE and MAE in the table are multiplied by 100. The format is: mean $\pm$ std.
	\end{tablenotes}
	\label{ablation_study}
\end{table*}

It is observed from Table~\ref{ablation_study} that model-1 performs the worst estimation results. Compared with the model-1, the maximum reduction of RMSE, MAE, and score for the SKDAN model are 49.6\%, 48.6\%, and 47.6\%, respectively. This indicates that the multi-head self-attention mechanism is the most important part of the SKDAN model, as it can accurately capture the domain-invariant features in battery charge data collected in different conditions. The results of model-3 analysis suggest that it is necessary to employ the CNN to further refine the degradation-related features in the subspace, instead of directly mapping them to the battery SOH. Furthermore, although model-2 and model-4 yield smaller estimation errors, they are still inferior to the SKDAN model, implying that the distillation operation and the smoothness constraint contribute to enhancing the model's performance. Consequently, the ablation experiments demonstrate that systematically integrating the above  parts results in better estimation performance of the SKDAN model.

\subsection{Visualization of feature distributions}
\begin{figure}[!htb]
	\centering

	\subfigure[Voltage in 0\%-60\% SOC range]{\label{0_60_raw}
		\includegraphics[width=0.32\textwidth]{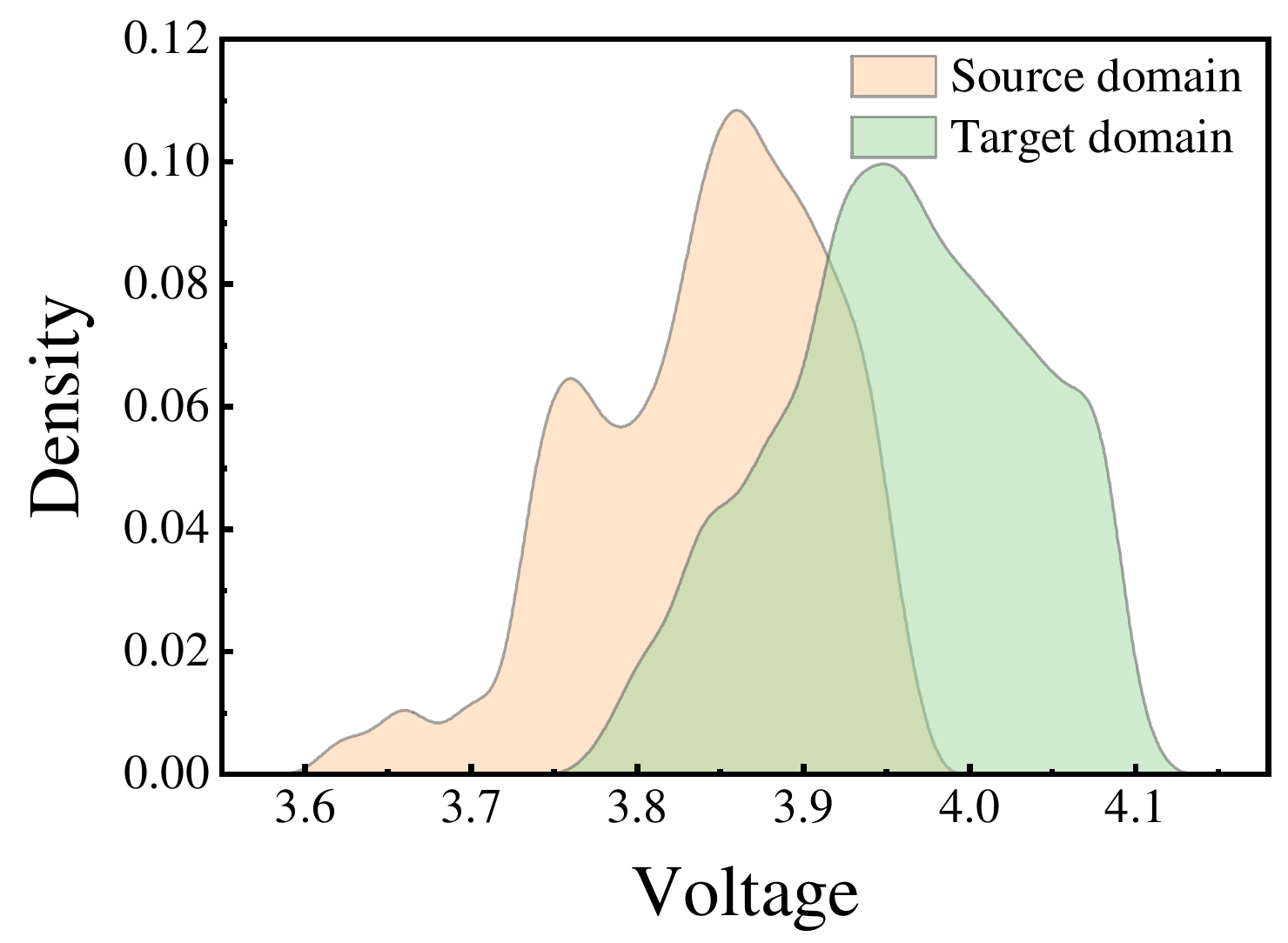}}
	\subfigure[Voltage in 20\%-80\% SOC range]{\label{20_80_raw}
		\includegraphics[width=0.32\textwidth]{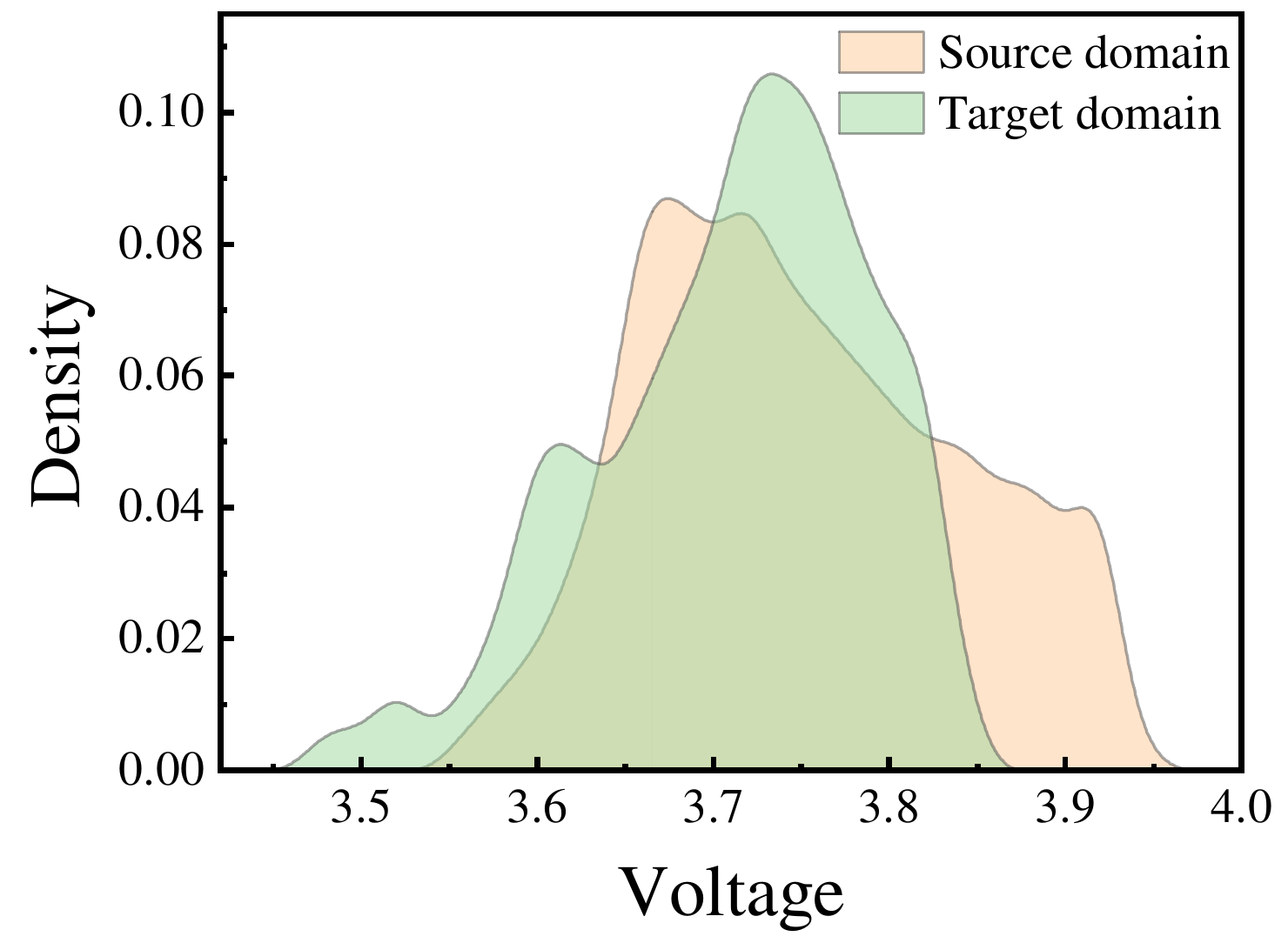}}
	\subfigure[Voltage in 40\%-100\% SOC range]{\label{40_100_raw}
		\includegraphics[width=0.32\textwidth]{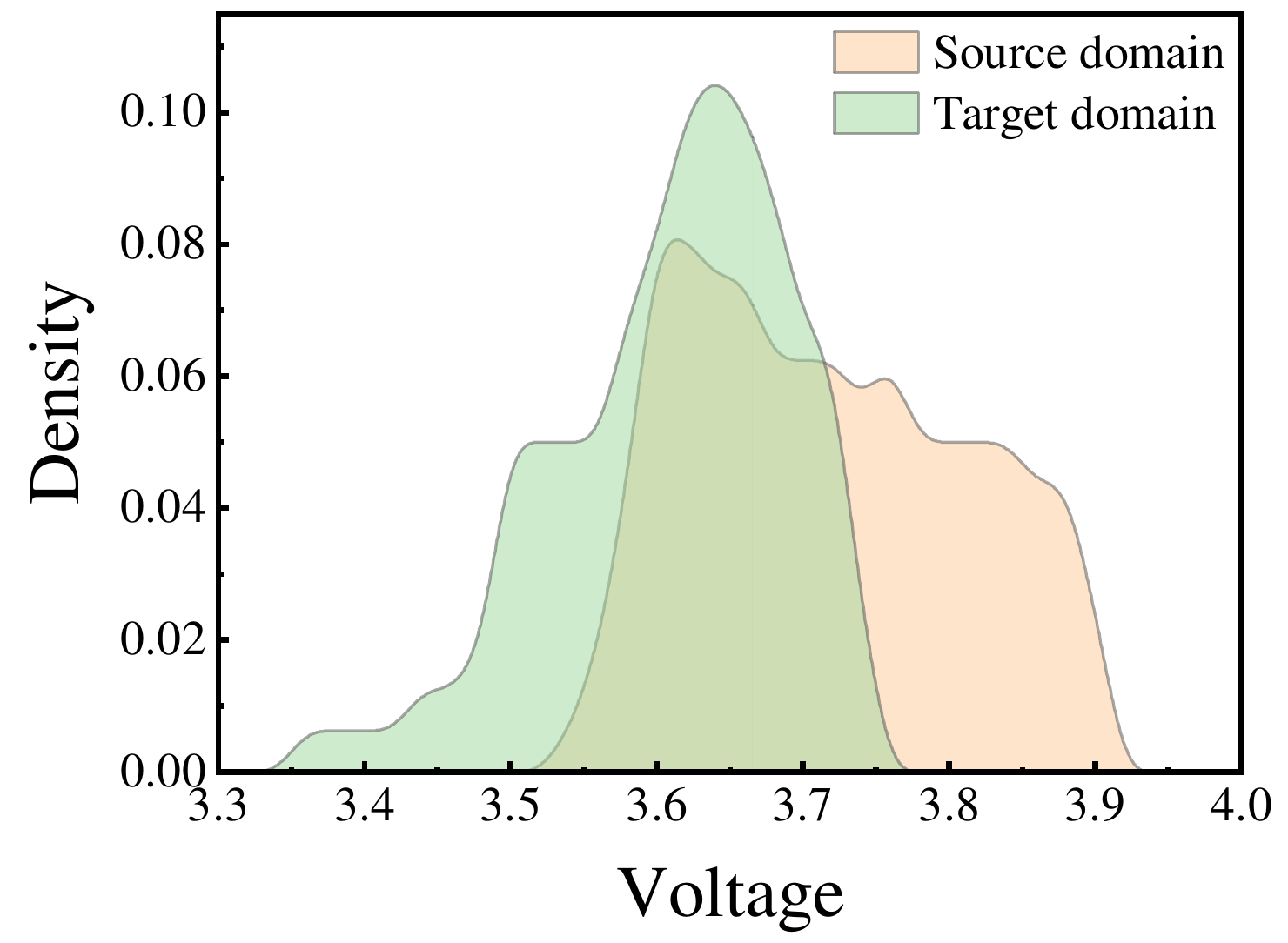}}\\
	
	\subfigure[Features in 0\%-60\% SOC range]{\label{0_60_features}
		\includegraphics[width=0.32\textwidth]{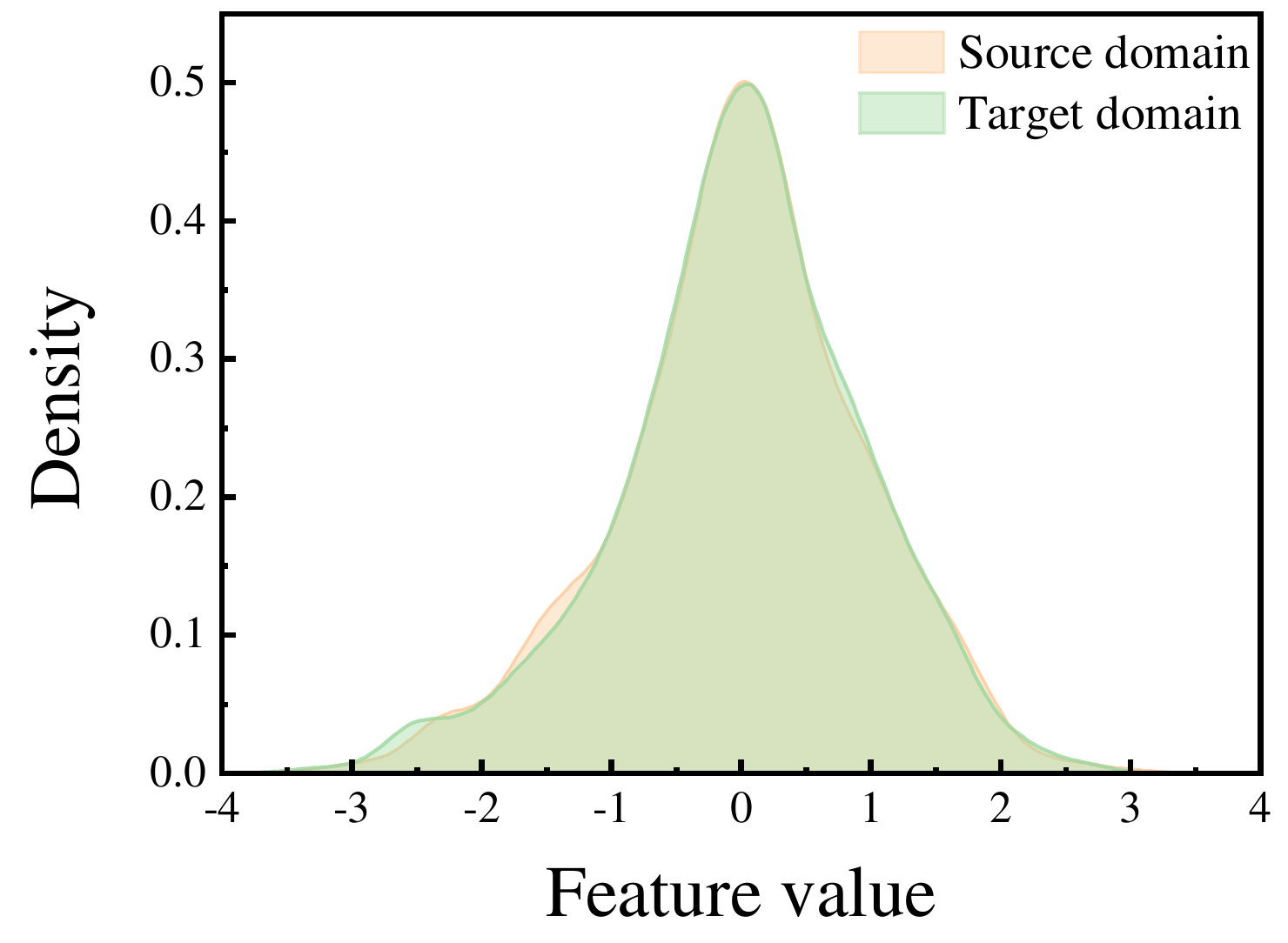}}
	\subfigure[Features in 20\%-80\% SOC range]{\label{20_80_features}
		\includegraphics[width=0.32\textwidth]{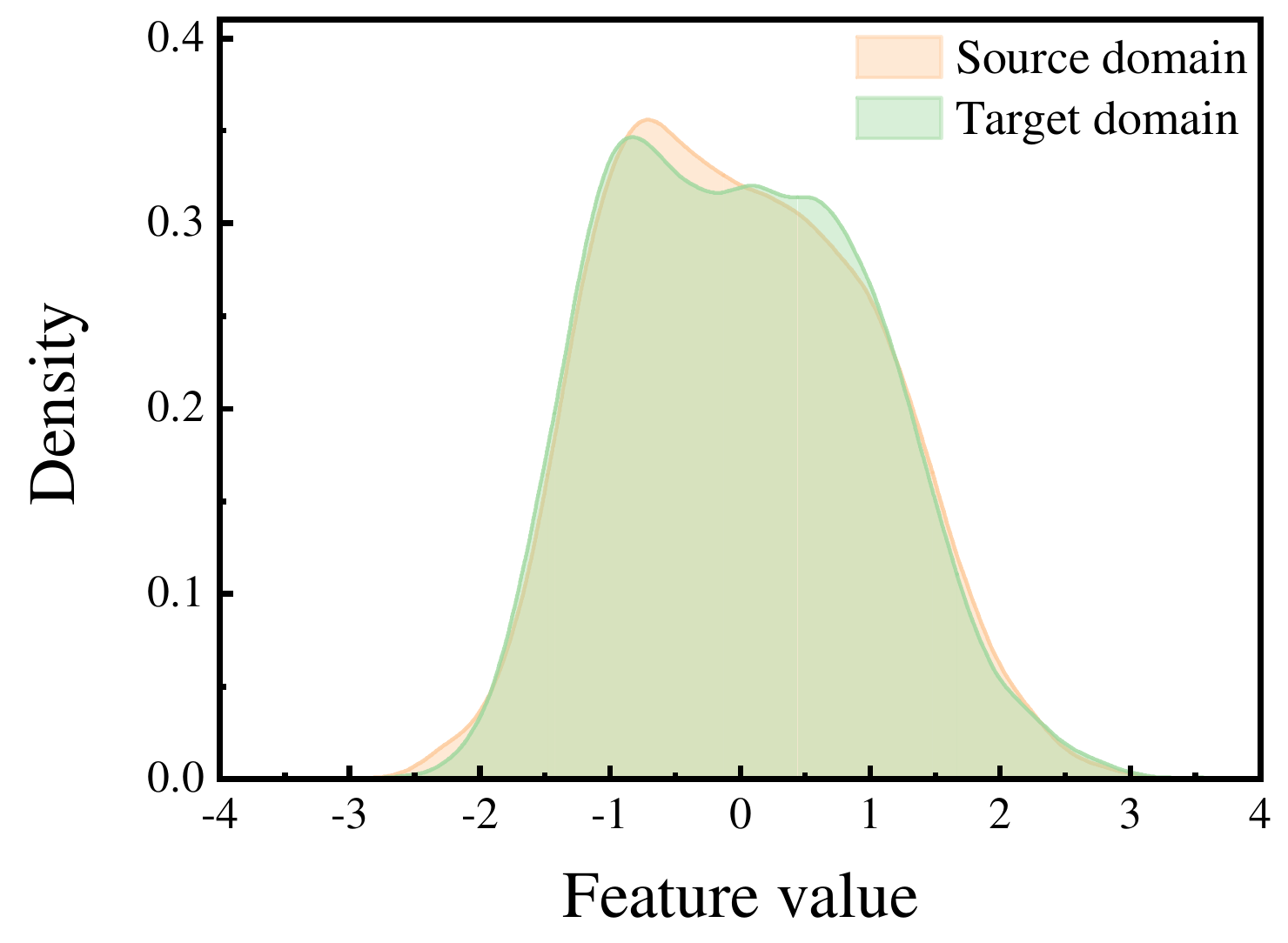}}
	\subfigure[Features in 40\%-100\% SOC range]{\label{40_100_features}
		\includegraphics[width=0.32\textwidth]{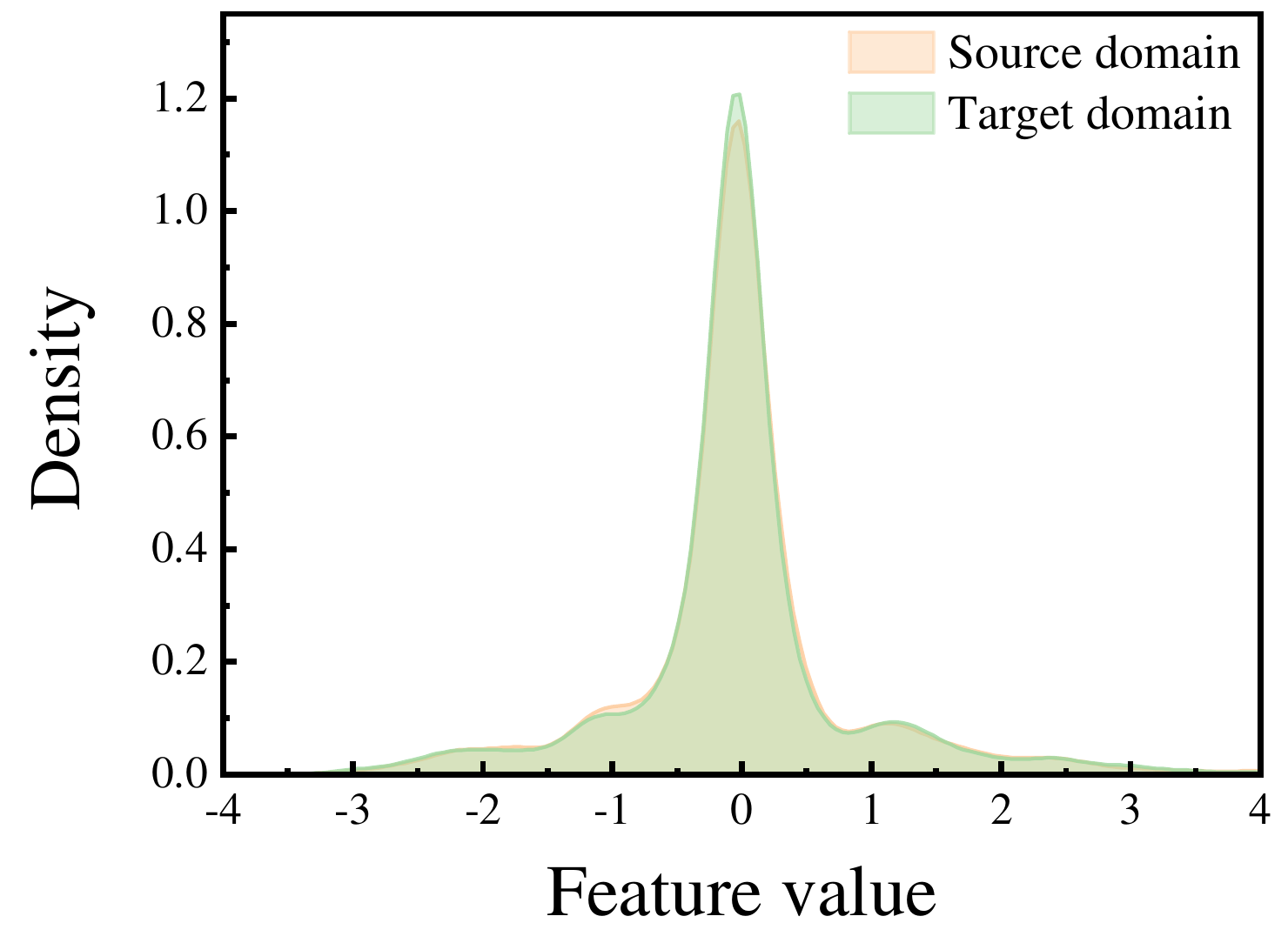}}
	\caption{The probability density distribution of the original data and extracted features}
	\label{distribution} %% label for entire figure
\end{figure}

The SKDAN model automatically extracts degradation-related features from the charge curves and maps them to the battery SOH. To investigate the effect of the SAD module and MK-MMD minimization on feature extraction, we conducted a visual analysis of the charge curves and degradation-related features from the transfer experiments in the CALCE dataset. The kernel density estimation is applied to calculate the probability density distribution of the original signal (here the voltage signal is chosen) and the degradation-related features, as shown in Figs.~\ref{0_60_raw}-\ref{40_100_features}.

Figs.~\ref{0_60_raw}-\ref{40_100_raw} clearly show that there is a significant difference in the voltage signal in the same SOC range for shallow and full cycles, as the DOD and SOC range affect the internal reaction of the battery. This domain discrepancy poses a challenge to capacity estimation at shallow-cycle conditions, making it infeasible to directly apply the learned knowledge from the source domain to the target domain. Thanks to the SAD feature extractor and MK-MMD minimization, the degradation-related features in different domains show a similar distribution in the subspace from Figs.~\ref{0_60_features}-\ref{40_100_features}. This suggests that the model can effectively extract domain-invariant features and reduce the effect of domain discrepancy, thereby significantly improving the model's performance.

\section{Conclusion}\label{conclusion}
Accurately estimating the SOH of a battery operating under shallow-cycle conditions is a challenging task. Herein, we develop a non-invasive and online UDTL-based SKDAN method to estimate the SOH of the shallow-cycle battery. The SKDAN model leverages knowledge learned from information-rich charge data in the source domain and transfers it to the target domain, which is robust against domain discrepancy. The performance of the SKDAN model is assessed on various transfer experiments in the CALCE and SNL datasets. In the single-domain knowledge transfer, the SKDAN model shows superior performance on different SOC ranges within an RMSE of 2\%, MAE of 1.7\%, and a score of 0.45. The comparative experiments demonstrate that the SKDAN model has a better capacity for extracting domain-invariant features and achieves a smaller estimation error than other models. The SKDAN model can capture the common degradation characteristics for knowledge transfer in different operating conditions to improve estimation performance, regardless of battery temperature and discharge rate. Importantly, the SKDAN model achieves a similar estimation to single-domain transfer for cross-domain transfer tasks without prior knowledge of manufacturers, chemical materials, and degradation mechanisms. This provides an efficient way to decrease the time and cost of developing data-driven SOH prediction models for newly manufactured batteries. In addition, the ablation studies and feature visualization further verify the effectiveness of the proposed model. This work highlights the promise of combining UDTL with charge curves to estimate SOH for shallow-cycle batteries. In the future, the SKDAN method can be applied to more complex operating scenarios, including more complicated charge and discharge protocols and the estimation of SOH for shallow-cycle batteries based on flexible SOC range charge data.

\section*{Acknowledgments}
Xin Chen acknowledges the funding support from the National Natural Science Foundation of China under grant No. 21773182 and the support of HPC Platform, Xi’an Jiaotong University.

\bibliography{cas-refs}
\end{document}